\documentclass[twoside,twocolumn,9pt]{article}
\usepackage{extsizes}
\usepackage[super,sort&compress,comma]{natbib}
\usepackage[version=3]{mhchem}
\usepackage[left=1.5cm,right=1.5cm,top=1.785cm,bottom=2.0cm]{geometry}
\usepackage{balance}
\usepackage{mathptmx}
\usepackage{sectsty}
\usepackage{graphicx}
\usepackage{lastpage}
\usepackage[format=plain,justification=justified,singlelinecheck=false,
  font={stretch=1.125,small,sf},labelfont=bf,labelsep=space]{caption}
\usepackage{float}
\usepackage{fancyhdr}
\usepackage{fnpos}
\usepackage[english]{babel}
\usepackage{array}
\usepackage{droidsans}
\usepackage{charter}
\usepackage[T1]{fontenc}
\usepackage[utf8]{inputenc}
\usepackage[usenames,dvipsnames]{xcolor}
\usepackage{setspace}
\usepackage[compact]{titlesec}
\usepackage{hyperref}
\usepackage{epstopdf}

\usepackage{placeins}
\usepackage{booktabs}
\usepackage{adjustbox}
\usepackage{tabularx}
\usepackage{pdflscape}
\usepackage{longtable}
\usepackage{soul}
\usepackage{hypcap}
\usepackage{subcaption}
\usepackage{pifont}
\usepackage{threeparttable}
\usepackage{ragged2e}
\usepackage{amsmath}
\usepackage{amssymb}
\usepackage{textcomp}
\usepackage{iftex}
\usepackage{gensymb}

\definecolor{cream}{RGB}{222,217,201}

\addto{\captionsenglish}{%
  
}
\providecommand{\Ca}{C\ensuremath{\alpha}}

\ifPDFTeX
  \DeclareUnicodeCharacter{03BC}{\ensuremath{\mu}}
  \DeclareUnicodeCharacter{202F}{\,}
\fi

\usepackage{listings}
\definecolor{promptText}{RGB}{25,25,25}
\definecolor{p53frame}{RGB}{31,119,180}\definecolor{p53bg}{RGB}{233,242,250}
\definecolor{memframe}{RGB}{44,140,60}\definecolor{membg}{RGB}{233,245,236}
\definecolor{ligframe}{RGB}{214,120,20}\definecolor{ligbg}{RGB}{252,244,232}
\definecolor{watframe}{RGB}{120,90,180}\definecolor{watbg}{RGB}{242,238,249}
\definecolor{ppframe}{RGB}{200,40,50}\definecolor{ppbg}{RGB}{251,236,237}
\lstdefinestyle{prompt}{%
  basicstyle=\footnotesize\ttfamily\color{promptText},
  breaklines=true,breakatwhitespace=false,columns=fullflexible,
  keepspaces=true,frame=single,framerule=1pt,framesep=8pt,
  xleftmargin=10pt,xrightmargin=4pt,aboveskip=6pt,belowskip=10pt,
  postbreak=\mbox{\textcolor{gray}{$\hookrightarrow$}\space},
}

\begin{document}

% RSC page and heading setup.
\pagestyle{fancy}
\thispagestyle{plain}
\fancypagestyle{plain}{\renewcommand{\headrulewidth}{0pt}}
\makeFNbottom
\makeatletter
\renewcommand\LARGE{\@setfontsize\LARGE{15pt}{17}}
\renewcommand\Large{\@setfontsize\Large{12pt}{14}}
\renewcommand\large{\@setfontsize\large{10pt}{12}}
\renewcommand\footnotesize{\@setfontsize\footnotesize{7pt}{10}}
\renewcommand\@biblabel[1]{#1}
\renewcommand\@makefntext[1]{\noindent\makebox[0pt][r]{\@thefnmark\,}#1}
\makeatother
\renewcommand{\thefootnote}{\fnsymbol{footnote}}
\renewcommand\footnoterule{\vspace*{1pt}%
\color{cream}\hrule width 3.5in height 0.4pt \color{black}\vspace*{5pt}}
\setcounter{secnumdepth}{0}
\renewcommand{\figurename}{\small{Fig.}~}
\sectionfont{\sffamily\Large}
\subsectionfont{\normalsize}
\subsubsectionfont{\bf}
\setstretch{1.125}
\setlength{\skip\footins}{0.8cm}
\setlength{\footnotesep}{0.25cm}
\setlength{\jot}{10pt}
\titlespacing*{\section}{0pt}{4pt}{4pt}
\titlespacing*{\subsection}{0pt}{15pt}{1pt}
\fancyfoot{}
\fancyfoot[LO,RE]{\vspace{-7.1pt}\includegraphics[height=9pt]{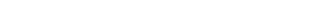}}
\fancyfoot[CO]{\vspace{-7.1pt}\hspace{13.2cm}\includegraphics{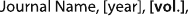}}
\fancyfoot[CE]{\vspace{-7.2pt}\hspace{-14.2cm}\includegraphics{head_foot/RF}}
\fancyfoot[RO]{\footnotesize{\sffamily{1--\pageref{MainArticleLastPage} ~\textbar  \hspace{2pt}\thepage}}}
\fancyfoot[LE]{\footnotesize{\sffamily{\thepage~\textbar\hspace{3.45cm} 1--\pageref{MainArticleLastPage}}}}
\fancyhead{}
\renewcommand{\headrulewidth}{0pt}
\renewcommand{\footrulewidth}{0pt}
\setlength{\arrayrulewidth}{1pt}
\setlength{\columnsep}{6.5mm}
\setlength{\bibsep}{1pt}

\makeatletter
\newlength{\figrulesep}
\setlength{\figrulesep}{0.5\textfloatsep}

\newcommand{\topfigrule}{\vspace*{-1pt}%
\noindent{\color{cream}\rule[-\figrulesep]{\columnwidth}{1.5pt}}}
\newcommand{\botfigrule}{\vspace*{-2pt}%
\noindent{\color{cream}\rule[\figrulesep]{\columnwidth}{1.5pt}}}
\newcommand{\dblfigrule}{\vspace*{-1pt}%
\noindent{\color{cream}\rule[-\figrulesep]{\textwidth}{1.5pt}}}
\makeatother

% RSC-style title, authors, and abstract.
\makeatletter
\twocolumn[
  \begin{@twocolumnfalse}
  {\includegraphics[height=30pt]{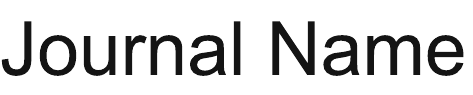}\hfill
  \raisebox{0pt}[0pt][0pt]{\includegraphics[height=55pt]{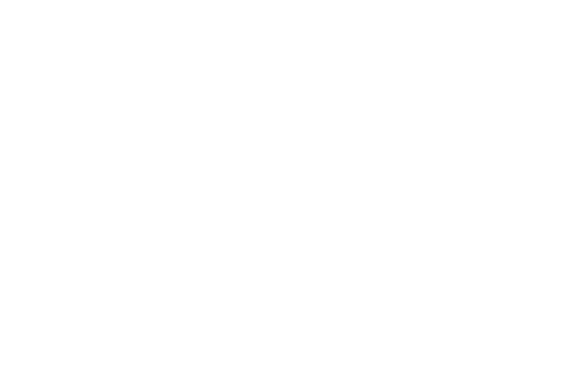}}\\[1ex]
  \includegraphics[width=18.5cm]{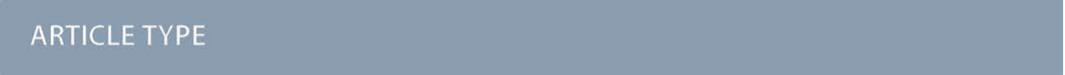}}\par
  \vspace{1em}
  \sffamily
  \begin{tabular}{m{4.5cm} p{13.5cm}}
  \includegraphics{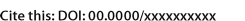}
  & \noindent\LARGE{\textbf{Automating MD simulations for Proteins using Large language Models: NAMD-Agent}} \\
  \vspace{0.3cm} & \vspace{0.3cm} \\
  & \noindent\large{Omid Barati Farimani,\textit{$^{a}$}$^{\ddag}$ Achuth Chandrasekhar,\textit{$^{a}$}$^{\ddag}$ and Amir Barati Farimani\textit{$^{a,b,c,d}$}$^{\ast}$} \\
  \includegraphics{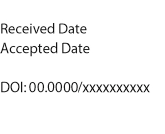} & \noindent\normalsize{
Molecular dynamics (MD) simulations are essential for understanding protein structure, dynamics, and function, but preparing, running, and analyzing simulations remains time-consuming and error-prone. We present an automated pipeline that combines large language model (LLM) agents with Python scripting and HTMD MCP tools to generate simulation-ready inputs for NAMD3/CHARMM, execute simulations, analyze outputs, and recover from build or runtime failures. The framework was evaluated across five biomolecular system classes: a protein-DNA complex (p53 DNA-binding domain bound to its response element), a protein-membrane system (M2 muscarinic receptor with iperoxo in a POPC/cholesterol bilayer), a protein-ligand series (five congeneric TYK2 inhibitors), a protein-water reference (ubiquitin), and a protein-protein complex (barnase-barstar). For the protein-DNA system, the automated workflow reproduced key metrics from an independent published benchmark. To distinguish framework performance from model-specific behavior, we repeated the complete protein-DNA study using three LLM orchestrators: Claude Opus-4.8, GPT-5.6 Sol, and Nemotron 3 Ultra. All three completed the workflow, but they differed in benchmark-ranking fidelity and by up to two orders of magnitude in token consumption and cost. Across all system classes, the agent recovered experimentally and computationally established behavior. Additional post-processing software was used to refine simulation outputs, enabling a complete and largely hands-free workflow. This approach reduces setup effort, limits manual errors, supports parallel handling of diverse biomolecular systems, and provides a robust, adaptable foundation for LLM-driven automation in computational structural biology.
  }
  \\
  \end{tabular}
  \end{@twocolumnfalse}
  \vspace{0.6cm}
]
\makeatother

\renewcommand*\rmdefault{bch}\normalfont\upshape
\rmfamily
\section*{}
\vspace{-1cm}
\footnotetext{\textit{$^{a}$~Department of Mechanical Engineering, Carnegie Mellon University, Pittsburgh, PA 15213, USA. E-mail: barati@cmu.edu}}
\footnotetext{\textit{$^{b}$~Department of Biomedical Engineering, Carnegie Mellon University, Pittsburgh, PA 15213, USA.}}
\footnotetext{\textit{$^{c}$~Department of Chemical Engineering, Carnegie Mellon University, Pittsburgh, PA 15213, USA.}}
\footnotetext{\textit{$^{d}$~Machine Learning Department, Carnegie Mellon University, Pittsburgh, PA 15213, USA.}}
\footnotetext{\ddag~These authors contributed equally to this work.}

\section{Introduction}
\label{sec:introduction}

Molecular dynamics (MD) has become a fundamental tool in structural biology, offering atomistic insight into biomolecular motion, stability, and interactions~\cite{hollingsworth2018molecular,huggins2019biomolecular}. Despite its broad impact, the path from a deposited structure to a reliable simulation trajectory remains complex and highly dependent on expert judgment. Preparing an MD system typically requires a sequence of connected steps, including cleaning and protonating the starting structure, selecting suitable force fields, parametrizing non-standard residues or ligands, constructing solvent boxes or membrane environments, assembling force-field-consistent topologies, defining minimization, equilibration, and production protocols, and finally designing system-specific analyses~\cite{hahn2022best,badar2022molecular, barati2024fast, barati2026accelerating}. Each step creates opportunities for subtle errors that may remain hidden but can strongly affect the reliability, reproducibility, and interpretation of the final simulation.

To reduce this burden, several automated MD platforms have been developed over the past decade. In the GROMACS ecosystem \cite{spoel2005gromacs}, CHAPERONg automates protein and protein--ligand MD workflows and provides a broad set of trajectory analyses, while Gmx-qk offers a lightweight bash- and GUI-assisted workflow for minimization, equilibration, production simulation, and MM/PBSA analysis~\cite{yekeen2023chaperong,singh2023gmx_qk}. Similar efforts have been introduced for other simulation environments, including easyAmber for Amber-based protein MD protocols, Admiral for docking-to-MD workflows in drug discovery, and PyAutoFEP for automated alchemical free-energy calculations~\cite{suplatov2020easyamber,baumgartner2020building,Karplus2002}. These tools have improved accessibility and helped standardize routine simulation procedures.
%However, many remain tied to predefined protocols, specific simulation engines, or fixed input templates, which can limit their flexibility when users need to adapt workflows across different classes of biomolecular systems.
High-Throughput Molecular Dynamics for Molecular Discovery (HTMD) improves these tools by providing a programmable and extensible platform that brings system setup, simulation execution, and analysis into a unified workspace~\cite{doerr2016htmd}.
%HTMD was designed to simplify the generation of simulation data while improving reproducibility across projects. Its guided builders support critical preparation steps such as refining input structures, assigning force-field parameters, constructing membranes, modeling glycans, and placing ions. By automating much of the setup process for the NAMD simulation engine~\cite{phillips2005scalable}, HTMD reduces manual intervention, lowers the chance of human error, and promotes consistency in system construction~\cite{doerr2016htmd}. Nevertheless, even with powerful platforms such as HTMD, users must still know which tools to call, how to sequence them, how to handle system-specific requirements, and how to connect the resulting simulations to appropriate downstream analyses.
 %Its step-by-step workflow does more than simplify the assembly of demanding systems like membrane proteins and multi-component complexes: because each decision is documented along the way, researchers gain a clear record they can return to, reinforcing both reproducibility and methodological rigor.

In parallel, recent advances in artificial intelligence, particularly large language models (LLMs), offer a promising route to make scientific workflows more adaptive and user-centered. LLMs gained widespread attention after the release of ChatGPT in November 2022, but their foundation can be traced back to the transformer architecture introduced in ``Attention Is All You Need''~\cite{vaswani2017attention}. This architecture enabled models to efficiently learn contextual relationships from large-scale datasets and generate coherent, context-aware responses. As a result, LLMs are increasingly being explored as tools for scientific reasoning, code generation, workflow planning, and human--computer interaction~\cite{mctear2022conversational,yang2024automatic,van2024adapted,nllb2024scaling,durante2024agent,chandrasekhar2025nanogpt}. For MD simulations, these capabilities are especially attractive because they can help translate high-level scientific intent into concrete computational actions.

Recent LLM-based and agentic simulation frameworks have begun to demonstrate this potential. Instead of simply executing predefined scripts, LLM agents can interpret natural-language instructions, select tools, generate code or input files, monitor execution, and adapt to intermediate errors~\cite{shi2026mdagent2, chandrasekhar2026material}. {Three of these systems make particularly useful points of comparison
  for the framework described here. DynaMate couples a planner, an MD module, and an
  analyzer through a LiteLLM loop to run protein and protein--ligand simulations in
  GROMACS with AMBER ff14SB and GAFF2 ligand parameters, and reports MM/PB(GB)SA
  binding free energies across twelve benchmark systems~\cite{guilbert2025dynamate}.
  MDCrow drives more than forty tools from a LangChain chain-of-thought agent over
  OpenMM, with literature and database retrieval built in, and is scored by expert
  human grading over twenty-five tasks~\cite{campbell2025mdcrow}. MDAgent assigns
  four role agents to a case-based skill and memory store and spans the widest
  methodological range of the three, including umbrella sampling, replica exchange,
  and coarse-grained Martini simulations in GROMACS~\cite{ma2026mdagent}. Each,
  however, binds its agent to a single simulation engine and tool implementation,
  and each is judged chiefly on whether the workflow completes rather than on
  whether the resulting physics matches an independent published result.} 
%MDAgent2 emphasized MD-specific question answering and simulation-code generation, particularly for LAMMPS-based workflows~\cite{shi2026mdagent2}.
PolyJarvis connected natural-language input with MCP-based tool orchestration and the RadonPy platform for autonomous polymer MD simulations~\cite{zhao2026polyjarvis}. More broadly, MatSciAgent and other multi-agent atomistic-simulation frameworks have shown that LLMs can coordinate materials retrieval, structure generation, simulation setup, execution, and post-processing across diverse computational tasks~\cite{chaudhari2026matsciagent,vriza2026multiagent}. Benchmarking efforts such as MDGYM further emphasize the need to evaluate AI agents on realistic molecular-simulation tasks rather than isolated demonstrations~\cite{kumar2026mdgym}.

A key requirement for reliable agentic simulation is the connection between LLM reasoning and structured, domain-specific computational tools. The Model Context Protocol (MCP) provides such a bridge by allowing LLMs to interact with external software environments through controlled and reproducible interfaces~\cite{hou2025model}. Through MCP, an LLM can access tools, retrieve structured data, and execute domain-specific operations while preserving context across workflow steps. For MD, this means that complex procedures can be decomposed into modular and verifiable actions, such as structure preparation, parameter assignment, system building, simulation setup, and trajectory analysis. Each action can then be executed through well-defined tool calls rather than informal free-text instructions~\cite{zhao2026polyjarvis,kang2026enzyseek}. This design improves traceability, reduces ambiguity, and helps preserve the deterministic operations required for trustworthy simulation workflows.

We asked a complementary question: \textbf{can an MD agent be made both broad and self-validating?}\\
In this work, we present an agentic framework~\cite{jadhav2024llm,ock2024adsorb,raman2025llm,chaudhari2025modular,zeng2025llm} that couples an LLM to the HTMD molecular-dynamics toolkit through an MCP server. The framework allows high-level scientific intent to be translated automatically into validated build, simulation, and analysis workflows. Rather than evaluating the system on a single benchmark, we demonstrate its generality across five representative case studies spanning major classes of biomolecular simulation: (i) a protein--DNA complex, in which the p53 DNA-binding domain and its R248 oncogenic mutants are built, simulated, and compared to quantify how the mutation perturbs DNA recognition (ii) protein--protein interfaces, including the barnase--barstar complex and 1AY7, to recover interface stability and binding determinants (iii) protein--ligand systems, where ligand parametrization, docking-pose preparation, and binding analysis are orchestrated end to end (iv) membrane-embedded proteins, which exercise automated bilayer construction and equilibration and (v) solvated single-protein dynamics as a baseline for folding and stability assessment. Across all five case studies, the agent autonomously selects and sequences the appropriate HTMD tools, including structure preparation, parametrization, minimization, equilibration, production simulation, and downstream RMSD/RMSF, hydrogen-bond, and interface analyses. In doing so, the framework combines the flexibility of natural-language interaction with the reproducibility of deterministic MD operations, providing a practical step toward more accessible, reliable, and generalizable biomolecular simulation workflows.

\section{Methods}
\label{sec:methods}

\subsection{Model: LLM Backbone}

The reasoning core of the framework is an LLM operated in an agentic, tool-using configuration. We employed Claude (Anthropic) opus 4.8 as the backbone model \cite{anthropic_claude_code}, accessed through the Claude Code agentic CLI (command-line interface), which exposes the model's function-calling capability. In this mode, the model does not merely generate text, instead it operates in a closed perception–action loop:
Given a natural-language objective and the descriptions of the available tools, it (i) reasons about the current state of the task, (ii) selects an appropriate tool, (iii) emits a structured call
specifying the tool name and its arguments, (iv) receives the tool's output as an observation, and (v) repeats until the objective is satisfied. This loop allows multi-step molecular dynamics
workflows each comprising dozens of interdependent operations to be planned and executed autonomously, with the model conditioning each decision on the results of preceding steps rather than following
a fixed script.

The model serves three distinct roles within the pipeline. First, as a planner, it decomposes a high-level request (e.g., "build, equilibrate, and analyze the p53–DNA complex") into an ordered sequence
of concrete operations. Second, as an orchestrator, it maps each operation onto the correct simulation tool and supplies physically and chemically appropriate parameters. Third, as an interpreter, it
inspects intermediate outputs' structure-quality reports, energy traces, convergence metrics and adapts subsequent steps accordingly, including error recovery when a step fails.

  %No fine-tuning wasperformed; the model was used in a zero-/few-shot setting, with domain behavior governed entirely by the tool schemas and system-level instructions [specify your system prompt / context if applicable].[Optionally state: model version, context-window size, temperature/decoding settings, and any deterministic-execution constraints used for reproducibility.]

\subsubsection{MCP Layer}
Communication between the LLM and the simulation software was mediated by the MCP, an open client–server standard for connecting language models to external tools, data, and resources. In our architecture, the LLM runtime acts as the MCP client (host), while a purpose-built MCP server ("htmd-mcp") wraps the HTMD/NAMD simulation and analysis stack and exposes its functionality as a catalog of callable tools. Client and server communicate over a [stdio / local transport] channel using structured JSON-RPC messages. At session start, the client queries the server, which advertises each tool together with a machine-readable schema describing its name, purpose, and typed input arguments. This schema-driven design is what allows the model to discover and correctly
invoke operations that it was never explicitly trained on.

The HTMD-MCP server registers 45 tools spanning the complete simulation lifecycle, organized into functional groups.
%: structure retrieval and inspection (e.g., fetch_pdb, inspect_structure,clean_structure); system construction (build_protein_waterbox, build_protein_nucleic, build_protein_ligand, build_protein_membrane, build_protein_protein_complex); preparation and parametrization (prepare_protein, parametrize_ligand); simulation setup and execution on the NAMD engine (setup_minimization, setup_equilibration, setup_production, run_namd3, monitor_namd_run); quality assurance(bond_length_sanity, quality_assurance_report); and downstream trajectory analysis (compute_rmsd, compute_rmsf, analyze_hbonds, interface_contacts, free_energy_landscape, mm_pbsa, and others). 
When the model issues a tool call, the MCP server validates the arguments against the tool's schema, executes the corresponding HTMD/NAMD routine in the underlying conda environment, and returns the
result file paths, numerical metrics, or status reports to the model as an observation. Crucially, each tool encapsulates a deterministic, version-controlled operation; the MCP layer therefore
preserves the reproducibility and methodological rigor of the established HTMD workflow while exposing it to natural-language control, cleanly separating the model's probabilistic reasoning from the simulation's deterministic execution.

\subsection{Agentic Workflow}

\begin{figure*}[hbt!]
\centering
\includegraphics[width=\textwidth]{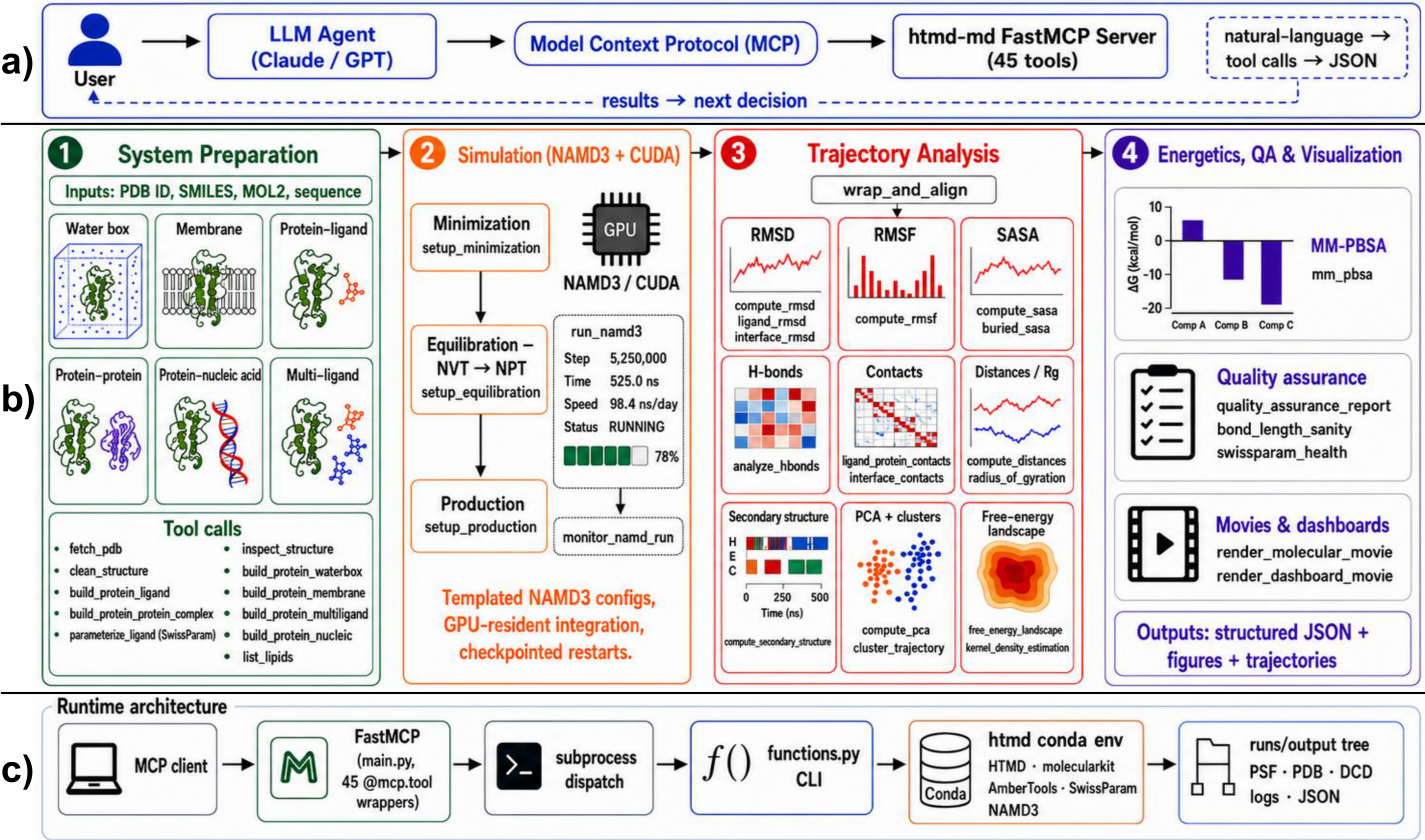}
\caption{\footnotesize
\textbf{(a)} Architecture and workflow of htmd-md for agentic protein molecular dynamics.
The htmd-md framework connects a user-facing LLM agent to a FastMCP server through the
Model Context Protocol, enabling natural-language requests to be translated into 45
structured molecular-dynamics tool calls.
\textbf{(b)} The workflow supports end-to-end protein MD tasks, including system
preparation, GPU-accelerated NAMD3 simulation, trajectory analysis, energetics, quality
assurance, and visualization. Major functions include building water, membrane,
protein--ligand, protein--protein, protein--nucleic acid, and multi-ligand systems,
followed by minimization, equilibration, production, and analyses such as RMSD, RMSF,
SASA, hydrogen bonds, contacts, PCA, and free-energy landscapes. The dashed feedback
loop highlights the agentic design, where simulation results can guide subsequent
LLM-driven decisions and generate organized outputs such as JSON files, figures,
trajectories, logs, PSF, PDB, and DCD files.
\textbf{(c)} The htmd-md runtime architecture connects the MCP client to FastMCP tool
wrappers, command-line functions, and the htmd conda environment to run molecular
dynamics tasks. All outputs are saved in an organized run folder, including structures,
trajectories, logs, figures, and JSON results.
}
\label{fig:NAMD_Agent_Flowchart}
\end{figure*}

% \subsection{Agentic workflow}

In this work, we developed htmd-md, an MCP-based server that connects a user-facing LLM agent to an automated protein molecular dynamics workflow through the Model Context Protocol (Figure \ref{fig:NAMD_Agent_Flowchart}a). The framework exposes 45 LLM-callable tools that translate natural-language instructions into structured tool calls for system preparation, simulation, trajectory analysis, energetics, quality assurance, and visualization. The system preparation module supports multiple biomolecular setups, including solvated proteins, membrane systems, protein–ligand complexes, protein–protein complexes, protein–nucleic acid systems, and multi-ligand systems, using inputs such as PDB IDs, SMILES strings, MOL2 files, or sequences .\\ Prepared systems are then simulated through automated NAMD3/CUDA workflows involving minimization, NVT/NPT equilibration, production runs, checkpointing, and run monitoring. After simulation, trajectories are processed and analyzed using standard MD observables, including RMSD, RMSF, SASA, hydrogen bonds, molecular contacts, distances, radius of gyration, secondary-structure evolution, PCA-based clustering, and free-energy landscape estimation. The framework also includes MM-PBSA energetic analysis, quality-control checks, molecular movie generation, dashboard rendering, and structured output organization, producing reproducible files such as JSON reports, figures, logs, PSF/PDB structures, and DCD trajectories. The returned results can be passed back to the LLM agent, enabling an iterative decision-making loop in which each simulation or analysis result informs the next automated action ({Figure~\ref{fig:NAMD_Agent_Flowchart}b}).
The Runtime architecture explains how the software is implemented behind the scenes. The workflow starts with an MCP client, which communicates with a FastMCP layer containing the main Python entry point and the 45 "@mcp.tool" wrappers. These wrappers provide the interface between the LLM and the underlying scientific functions. Requests are then dispatched through a subprocess layer to the command-line logic implemented in functions.py/CLI. From there, the commands interact with the htmd conda environment, which includes the scientific software stack needed to perform the tasks, such as HTMD, moleculekit, AmberTools, SwissParam, and NAMD3. Finally, results are organized into a runs/output tree, where output files such as PSF, PDB, DCD, logs, and JSON are stored (Figure \ref{fig:NAMD_Agent_Flowchart}c).

\subsection{Systems and benchmarks}

The five systems and their independent references were: \textbf{protein--DNA}-p53
DNA-binding domain bound to a 12-bp response element (PDB 2AC0), benchmarked against
Balasundaram and Priya Doss\cite{balasundaram2022unraveling}.
\textbf{protein--membrane}-active M2 muscarinic receptor (PDB 4MQS) with the agonist
iperoxo in a POPC/cholesterol (9:1) bilayer, benchmarked against the established
M2--iperoxo structural pharmacology and a published membrane-MD tutorial workflow\cite{kruse2013activation}.
\textbf{protein--ligand} Full set (16) congeneric TYK2 inhibitors, benchmarked against the
Schr\"odinger FEP+ relative-binding-free-energy set and the corresponding experimental
affinities\cite{wang2015accurate}. \textbf{protein--protein} barnase-barstar
(PDB 1BRS), benchmarked against Martin \& Frezza et al.\cite{martin2022dynamical}, and \textbf{protein--water} ubiquitin (PDB 1UBQ), the canonical
globular-protein stability reference \cite{vijay1987structure,long2011silico}. Unlike the other four systems, the ubiquitin protein–water run was not set up to reproduce a specific published MD paper{.} Visualization of the evolving system is performed using molecular graphics programs such as VMD \cite{humphrey1996vmd}, enabling detailed inspection of structural changes, membrane dynamics, and solvent behavior. {All the five source data details are given in Table~S1 (Supplementary Information)}.

\subsection{NAMD-Agent Toolset}

\subsubsection{Preprocessing Tools} 
NAMD-Agent begins its workflow by applying structure preprocessing techniques to ensure high-quality simulation input. The system uses a protein repair module in PDBFixer \cite{pdbfixer} to automatically address missing heavy atoms, incomplete residues, and nonstandard residues in PDB files. This step ensures compatibility with subsequent modeling tools and significantly reduces failure points during simulation setup.

\subsubsection{Simulation setup and system construction}

Every system was assembled through the same staged pipeline, with the agent selecting
the builder appropriate to each system class and carrying the output of one step into
the next. Construction began by obtaining and auditing the starting coordinates: a structure was fetched from the PDB (or read locally) and inspected for its chains, segments, bound ligands, ions, crystallographic waters, and chain breaks. This audit dictated the cleanup that followed. The protein was then prepared, with protonation
states assigned at the target pH via PROPKA, hydrogen-bond networks optimized, and
missing atoms rebuilt. Extraneous species such as crystallization additives or
reference ligands were stripped, and segments were assigned so that \texttt{psfgen}
could later build the topology.
Small molecules entered through a parallel branch. A ligand supplied as a SMILES string, an SDF or MOL2 file, or a binding pose extracted from a larger structure was
first standardized into a parameter-ready mol2 with explicit hydrogens, three-dimensional
coordinates, and initial charges, then parametrized with SwissParam to yield CGenFF-style
CHARMM topology and parameters. The prepared protein and ligand were combined by the builder matching the system class: a plain water box for soluble proteins, a
protein--ligand builder that seats the ligand in its docked pose (or, when an unbinding
setup is wanted, far from the receptor), a bilayer builder that embeds the receptor in a
chosen lipid mixture, or dedicated protein--nucleic and protein--protein complex builders.
Each builder solvates the assembly in TIP3P water, neutralizes the net charge, and adds
salt to the target concentration before running the CHARMM build  \cite{jo2008charmm, wu2014charmm, lee2016charmm}, emitting a matched pair of PSF and PDB files.

From the finished system, the agent generated the NAMD3 inputs for each stage in turn: a restrained minimization, a restraint-ramped equilibration that inherits the minimized
coordinates and velocities, and a production run that continues from the equilibrated
state, or invoked a single convenience wrapper that chains all three under one
workspace. Because every setup step stages the required parameter streams and passes the
coordinate, velocity, and box files forward automatically, the same construction recipe
reproduces a runnable system for any of the five classes. Table~\ref{tab:tools_summary}
groups these tools into functional categories; the full per-tool descriptions are given in {Table~S2} (Supplementary Information).

\begin{table*}[htbp]
\centering
\caption{System-construction tool categories in the \texttt{htmd-mcp} server, in
pipeline order. Each category bundles several tools that chain automatically. The
complete per-tool listing is given in {Table~S2} (Supplementary
Information).}
\label{tab:tools_summary}
\small
\begin{tabular}{@{}p{3.2cm} p{11.2cm}@{}}
\toprule
Category & Role in the build pipeline \\
\midrule
Acquire \& inspect & Fetch a structure from the PDB or a local file and audit its chains, segids, ligands, ions, waters, and chain breaks. \\
\addlinespace
Prepare \& clean & Protonate the protein (PROPKA), optimize the H-bond network, rebuild missing atoms, strip crystallization additives, and assign segids for \texttt{psfgen}. \\
\addlinespace
Ligand parameters & Standardize any ligand input (SMILES/SDF/mol2/pose) to a parameter-ready mol2 and derive CGenFF/SwissParam CHARMM topology and parameters. \\
\addlinespace
Build solvated system & Assemble the class-specific system (water box, protein--ligand, multi-ligand, membrane, protein--nucleic, or protein--protein), then solvate in TIP3P, neutralize, and add salt. \\
\addlinespace
Generate run inputs & Write the chained NAMD3 inputs, restrained minimization, restraint-ramped equilibration, and production (optional HMR), or a one-shot wrapper. \\
\bottomrule
\end{tabular}
\end{table*}

\subsection{Common simulation protocol}

The simulations used CHARMM36m for proteins, CHARMM36 for lipids and nucleic acids, and CGenFF/SwissParam for ligands, together with the TIP3P water model. They employed PME electrostatics with a
12~\AA{} real-space cutoff (10~\AA{} switch), \texttt{rigidBonds} on hydrogens, a Langevin thermostat and Langevin piston barostat, and NAMD3 in GPU-resident mode on a
single NVIDIA RTX 2080 Ti (11~GB). Production lengths were 100~ns (protein--DNA,
protein--membrane), $\sim$300~ns (protein--protein), a multi-nanosecond reference run
(protein--water), and per-ligand production for the protein--ligand series. System-specific details and force-field streams are given with each result.

\subsection{Trajectory analysis and comparison}

Once a production trajectory was in hand, the agent post-processed it through a second
suite of tools and, crucially, distilled each run into quantities directly comparable to
its literature benchmark. Trajectories were first re-imaged and aligned to remove
periodic-boundary artifacts, after which global stability was characterized through
RMSD\cite{Kabsch1976}, per-residue RMSF\cite{Amadei1993,Seeliger2010}, radius of gyration\cite{Flory1969,Kohn2004}, and DSSP secondary structure, alongside
solvent exposure, inter-selection distances, and hydrogen-bond networks \cite{Baker1984,McDonald1994}. System-class observables followed: ligand heavy-atom RMSD, ligand--protein contact fingerprints, and
single-trajectory MM-GBSA for protein--ligand systems, CAPRI-style interface RMSD, interfacial contacts, and buried SASA \cite{Lee1971,Shrake1973} for complexes, and where informative, a
conformational-landscape branch built from Cartesian PCA, clustering, and a free-energy
landscape over the leading principal components. Quality-assurance and visualization
tools closed the loop checks on energy, temperature, pressure, and volume drift,
bond-length sanity, and trajectory movies while a dedicated comparison tool overlaid
metrics across systems, mutants, or replicas to produce the benchmark figures.
Table~\ref{tab:analysis_summary} groups these analyses into functional categories: the full per-tool descriptions are given in {Table~S3}.
(Supplementary Information).

\begin{table*}[htbp]
\centering
\caption{Trajectory-analysis and comparison tool categories in the
\texttt{htmd-mcp} server. Class-specific and landscape categories are applied only
where relevant; the complete per-tool listing is given in {Table~S3}
(Supplementary Information).}
\label{tab:analysis_summary}
\small
\begin{tabular}{@{}p{3.4cm} p{11.0cm}@{}}
\toprule
Category & What it measures \\
\midrule
Preprocess & Re-image waters/ions and align the trajectory to remove periodic-boundary artifacts. \\
\addlinespace
Global stability & Backbone RMSD, per-residue RMSF, radius of gyration, and DSSP secondary structure. \\
\addlinespace
Solvation \& contacts & Solvent-accessible surface area, inter-selection distances, and hydrogen-bond networks. \\
\addlinespace
Protein--ligand & Ligand pose drift, per-residue binding fingerprint, and single-trajectory MM-GBSA. \\
\addlinespace
Interfaces & CAPRI-style interface RMSD, per-residue interface contacts, and buried interfacial SASA. \\
\addlinespace
Conformational landscape & Cartesian PCA, clustering, free-energy landscape, and kernel-density estimation over collective variables. \\
\addlinespace
QA \& visualization & Energy/temperature/pressure/volume drift dashboard, bond-length sanity, and trajectory movies. \\
\addlinespace
Comparison & Overlay of RMSD/RMSF/$R_g$/H-bonds across systems, mutants, or replicas. \\
\bottomrule
\end{tabular}
\end{table*}

\newpage

\section{Results and Discussion}
We open the Results with the protein--DNA system because, among the five system classes, it offers the most informative test of a closed-loop MD agent and therefore anchors the interpretation that follows. Unlike the individual single-trajectory systems (ubiquitin, the membrane GPCR, and the protein complexes), the p53 study is comparative \emph{by construction}. The three variants: wild type and the cancer hotspot mutants R248W and R248G, share an identical build and protocol, so differences among them provide a controlled readout of mutation rather than setup. The expected progressive loss of p53--DNA binding also provides an unambiguous direction against which the pipeline can be evaluated. This system is the most directly commensurable with the reference study: Balasundaram and Priya Doss simulated the \emph{same} starting structure (PDB 2AC0) with the \emph{same} force field (CHARMM36), so the comparison primarily isolates engine, thermostat, and analysis-definition effects. Finally, p53 exercises the full analysis stack in concert: backbone RMSD, per-residue RMSF, radius of gyration, and the interfacial hydrogen-bond network. These observables must tell a \emph{mutually consistent} mechanistic story: a loosened DNA-binding loop, a weakened interface, and a preserved fold, for the result to be trustworthy. Because it combines an internal control, a like-for-like literature benchmark, clinically important biology, and several orthogonal metrics that cross-validate one another, the p53 R248 system is the natural case in which to demonstrate and interpret what the agentic pipeline reproduces and where and why it diverges from the reference.

\subsection{\textbf{Protein--DNA}: the p53 R248 hotspot weakens the DNA interface in the correct order.}

\textbf{System and comparison design:}
The p53 DNA-binding domain (DBD, residues 94--312) bound to a 12-bp response element was
built from PDB \textbf{2AC0} and simulated in the wild-type (WT) form and as the two
cancer hotspot substitutions \textbf{R248W} and \textbf{R248G}, each for 100~ns of
single-replica production (CHARMM36, NAMD3, 2~fs timestep, frames every 20~ps,
$\sim$5000 frames per variant). Arg248 sits in the L3 loop of the DBD and makes a direct
minor-groove contact with the DNA, so mutating it is expected to perturb the
protein--DNA interface without unfolding the domain, exactly the behavior we test for.
The benchmark is Balasundaram \& Priya Doss\cite{balasundaram2022unraveling}, who studied the \emph{same} starting structure and the \emph{same} CHARMM36
force field, but with a different engine (GROMACS), a weak-coupling Berendsen thermostat,
and \texttt{gmx hbond}. To make the two studies commensurable we report analyses under
\emph{paper-matched} settings: backbone (N,\,C\Ca,\,C) superposition onto the built
crystal structure, exclusion of the three terminal residues at each end, and a
GROMACS-style hydrogen-bond criterion (donor--acceptor $\le 3.5$~\AA, angle $\le
30$\degree, strictly intermolecular protein$\leftrightarrow$nucleic pairs, minimum-image,
complex re-imaged on the protein). Table~\ref{tab:p53} collects the headline numbers
against the reference. Figure~\ref{fig:figure2} presents the per-variant time series and profiles.

\begin{table*}[htbp]
\centering
\caption{p53 DBD--DNA: our paper-matched observables (mean over the 100~ns production)
versus Balasundaram \& Priya Doss 2022. Both studies use PDB 2AC0 and CHARMM36; the
reference uses GROMACS/Berendsen/\texttt{gmx hbond}, whereas this work uses
NAMD3/Langevin/Wernet--Nilsson with the paper-matched criterion.
$\uparrow$/$\downarrow$ denote the direction of change relative to WT.}
\label{tab:p53}

% Increase table font, row height, and column spacing
\normalsize
\renewcommand{\arraystretch}{1.45}
\setlength{\tabcolsep}{11pt}

\begin{adjustbox}{max width=\textwidth}
\begin{tabular}{@{}l ccc c ccc@{}}
\toprule
& \multicolumn{3}{c}{\textbf{This work (NAMD3)}}
& & \multicolumn{3}{c}{\textbf{Reference (GROMACS)}} \\
\cmidrule(lr){2-4}
\cmidrule(lr){6-8}

\textbf{Quantity}
& \textbf{WT}
& \textbf{R248W}
& \textbf{R248G}
& 
& \textbf{WT}
& \textbf{R248W}
& \textbf{R248G} \\
\midrule

Protein--DNA H-bonds
& 8.24
& 7.83
& 7.15
&
& 9.21
& 8.54
& 6.77 \\

Backbone RMSD (\AA)
& 1.77
& 1.93
& 2.09
&
& $\sim$1.5
& 1.5--2.5
& 1.5--2.5 \\

Mean \Ca{} RMSF (\AA)
& 0.92
& 0.98
& 1.13
&
& \multicolumn{3}{c}{mutants $>$ WT (loops)} \\

Radius of gyration (\AA)
& 16.83
& 16.80
& 16.85
&
& 16.94
& 16.97
& 16.98 \\

\midrule

Interface trend (H-bonds)
& \multicolumn{3}{c}{WT $>$ R248W $>$ R248G $\downarrow$}
&
& \multicolumn{3}{c}{WT $>$ R248W $>$ R248G $\downarrow$} \\

\bottomrule
\end{tabular}
\end{adjustbox}
\end{table*}
\textbf{Global stability   backbone RMSD (Fig.~\ref{fig:figure2}a):}
All three trajectories equilibrate within the first $\sim$10~ns and then fluctuate about
a stable plateau, with no drift indicative of unfolding: the DBD scaffold is retained in
every variant. The plateau height, however, orders with mutational severity,
{WT $1.77 < $ R248W $1.93 < $ R248G $2.09$~\AA, reproducing the reference's
central structural observation that the mutants deviate further from the crystal pose
than wild type, and that the glycine substitution, which removes the Arg248 side chain entirely, deviates most. The absolute values sit slightly above the reference's
$\sim$1.5--2.5~\AA{} band, an offset we attribute below to engine/thermostat and to true
canonical (Langevin) sampling rather than Berendsen weak coupling. We flag one honest
caveat that the figure makes visible: under our \emph{default} analysis (fit to
production frame~0, termini included) the WT curve shows a late upward excursion after
$\sim$80~ns (last-20~ns mean RMSD $\approx$2.1~\AA), which can invert the WT/mutant
ordering. This is single-replica stochasticity for systems separated by $<1$~\AA, not a
genuine reversal. The paper-matched fit (to the crystal, termini excluded) removes the artifact and restores the expected ordering, and the agent's own report reached the same conclusion.

\begin{figure*}[!t]
\centering
\includegraphics[width=1\textwidth]{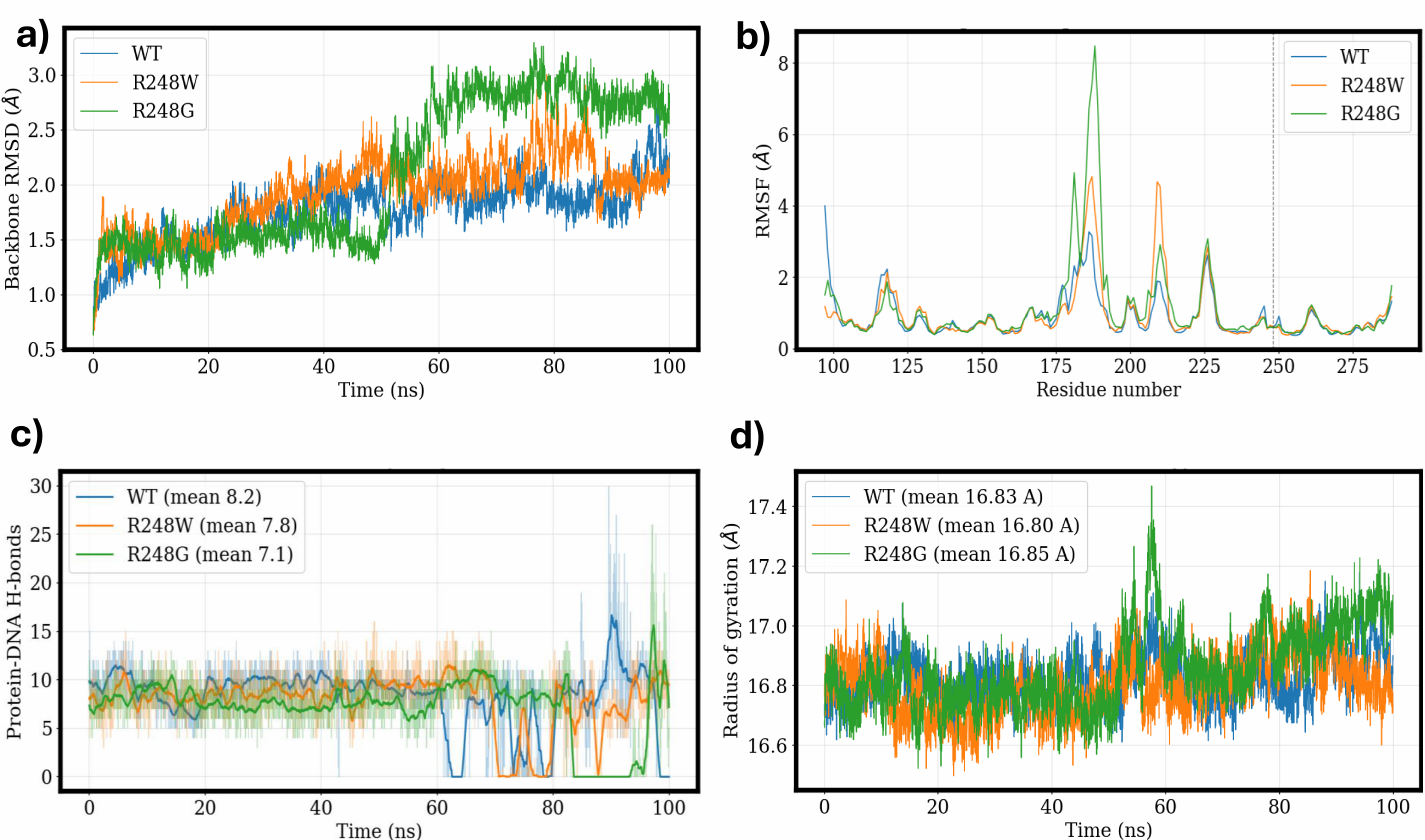}
\caption{\textbf{a)}Backbone \Ca{} RMSD versus time for p53 DBD--DNA WT, R248W, and R248G (100~ns, paper-matched fit to the crystal, termini excluded). All variants plateau without
unfolding; the mean RMSD orders WT $<$ R248W $<$ R248G, matching the benchmark's finding that the hotspot mutants deviate more than wild type. \textbf{b)}Per-residue \Ca{} RMSF for the three variants. Flexibility concentrates in the
L2/L3 DNA-binding loops ($\sim$181--189 and $\sim$209--210). R248G shows the largest loop fluctuations (up to $\sim$8.5~\AA{} near residues 186--189), R248W is intermediate, and WT is the most rigid, mirroring the mutational ordering seen in RMSD and H-bonds. \textbf{c)} Number of intermolecular protein--DNA hydrogen bonds versus time (paper-matched GROMACS-style criterion, $\le 3.5$~\AA/$\le 30$\degree). Time-averaged counts decrease WT $>$ R248W $>$ R248G (8.24 / 7.83 / 7.15), reproducing the benchmark's central result
that the R248 hotspot mutations progressively disrupt the p53--DNA interface, with R248G
the most disruptive. \textbf{(d)} Radius of gyration versus time. The nearly identical means (16.83/16.80/16.85~\AA{} for WT/R248W/R248G) show that the global compactness of the DBD is preserved.} 
\label{fig:figure2}
\end{figure*}

\textbf{Local flexibility per-residue RMSF (Fig.~\ref{fig:figure2}b):}
The fluctuation profiles are quiet across the folded $\beta$-sandwich core and rise
sharply at the same mobile loops in every variant, but with mutant-dependent amplitude.
The dominant peaks localize to the L2/L3 DNA-binding loops residues $\sim$181-189 and
$\sim$209--210, precisely the region carrying and flanking the Arg248 contact. R248G is
by far the most flexible: its 181--189 loop reaches $\sim$6.7--8.5~\AA{} of fluctuation
(residues 186-189), against $\sim$4.6--4.8~\AA{} for R248W and $\sim$3.3~\AA{} for WT at the same residues. This is the mechanistic reading of the H-bond and RMSD results:
removing the Arg248 side chain uncages the loop that anchors p53 on the DNA, so the loop
samples a wider ensemble and the interface loosens. The elevated loop mobility, and its
ordering WT $<$ R248W $<$ R248G, agrees with the reference's report of increased mutant
fluctuations in the loop regions ($\sim$165--172, $\sim$180--188, $\sim$208--215,
$\sim$240--250). The very large values at the extreme termini (residues $\sim$97--98 and
$\sim$288) reflect free-end tumbling and are excluded from the RMSD fit, they are not part of the interface signal.

\textbf{The interface itself protein--DNA hydrogen bonds (Fig.~\ref{fig:figure2}c):}
This is the decisive, most directly biological observable, and it is where the agentic pipeline reproduces the benchmark's headline result. The instantaneous protein--DNA H-bond count fluctuates over a broad range frame-to-frame (roughly 0--30 bonds), but its
time average falls monotonically with mutational severity: 8.24 (WT) $\rightarrow$
7.83 (R248W) $\rightarrow$ 7.15 (R248G). The reference reports the same ordering and the
same conclusion: WT $>$ R248W $>$ R248G, with R248G the most disruptive
(benchmark 9.21 / 8.54 / 6.77). Both studies therefore agree that mutating Arg248 weakens
the p53--DNA hydrogen-bond network and that the glycine substitution is the most damaging,
which is the biologically and clinically relevant finding (these are recurrent
loss-of-function hotspot mutations in human cancers). Our absolute counts run
$\sim$1--2~bonds below the reference; this is a hydrogen-bond definition effect, not a disagreement about the physics, because the trend and rank order are preserved.

\textbf{Compactness radius of gyration (Fig.~\ref{fig:figure2}d):}
$R_g$ is flat and essentially indistinguishable across the three variants
(16.83 / 16.80 / {16.85~\AA{}} for WT / R248W / R248G), confirming that the mutations
act \emph{locally} at the interface rather than by expanding or collapsing the fold and the
$\beta$-sandwich stays intact for all three. These values agree with the reference's
$R_g$ (WT/R248W/R248G $=$ 1.6939 / 1.6974 / 1.6975~nm, i.e.\ $\approx$16.94--16.98~\AA) to
within $\sim$0.15~\AA. We reproduce even the reference's subtle trend that the mutants are
marginally more expanded than WT, although at this magnitude ($<0.1$~\AA) the difference
is within single-replica noise and we do not over-interpret it.

\textbf{Discussion: what agrees, what differs, and why:}
Across every observable the \emph{biological} conclusion of the benchmark is reproduced:
R248 mutation weakens the p53--DNA interface (fewer interfacial H-bonds), R248G is the
most disruptive variant, the DNA-binding loops become progressively more mobile
(WT $<$ R248W $<$ R248G), and the overall fold and dimension are preserved. Four sources
account for the residual quantitative offsets, none of which overturn the trend.
\textbf{(1) Engine and thermostat.} The reference uses GROMACS with a Berendsen
weak-coupling thermostat, which damps fluctuations and keeps the system near its starting
structure. We use NAMD3 with a Langevin thermostat and Langevin-piston barostat, which
generate true canonical sampling and explore a slightly wider ensemble, biasing our
absolute RMSD upward. \textbf{(2) RMSD reference and termini.} Our default analysis used production frame~0 as the reference and retained the mobile N/C termini (which fluctuate by 4--8~\AA), thereby inflating the absolute RMSD and destabilizing the WT/mutant ordering. Fitting backbone atoms to the crystal structure with the termini excluded recovers lower magnitudes and the expected time-averaged ordering. \textbf{(3) Hydrogen-bond
definition.} \texttt{gmx hbond} counts a bond for donor--acceptor $\le 3.5$~\AA{} and angle
$\le 30$\degree; our detector applies the stricter Wernet--Nilsson distance-dependent
angular criterion, which returns $\sim$1--2 fewer bonds while preserving the trend, the
main reason our absolute counts sit below the reference. \textbf{(4) Single-replica
stochasticity.} Each variant is one 100~ns trajectory in both studies; for systems
separated by $<1$~\AA{} in RMSD the ranking is not statistically robust, which is exactly
why the WT default-analysis excursion could masquerade as a reversal. The honest remedy,
which we state as a limitation, is $\geq 3$ replicas per variant. The current single-replica design supports the qualitative mutational ranking (which is consistent across H-bonds, RMSF, and paper-matched RMSD) but not fine quantitative differences below the
$\sim$1~\AA/$\sim$1-bond level. In short, the agentic NAMD3 pipeline independently
recovers the benchmark's mechanism for p53 R248 loss of function, and the places where the numbers differ are traceable, expected, and transparently attributable to method choices
rather than to a failure of the simulation.
\subsection{\textbf{LLM comparison on Protein--DNA}}

{We evaluated the same three p53 DBD--DNA systems, which include the wild type and the R248W and R248G hotspot mutants under three different LLM
orchestrators: Claude Opus 4.8, GPT-5.6 Sol, and the open-weight Nemotron 3
Ultra. Each model received the same starting structure (PDB 2AC0) and the same
htmd-mcp tool surface, and then independently built the mutants, solvated,
minimised, equilibrated, ran 100~ns of production, and analysed the resulting
trajectories. Nothing else was shared between the three runs. Because the three
agents had each also chosen their own post-processing settings, all nine
trajectories were re-analysed from the raw coordinates under a single identical
protocol (paper-matched hydrogen-bond criterion, \texttt{name CA} backbone,
first and last three residues excluded, superposed on each system's built
structure, 5{,}000 frames at 0.02~ns), so that the columns of
Table~\ref{tab:p53_llm} differ only in the trajectories and not in how they were
measured.}

{Claude Opus 4.8 and GPT-5.6 Sol both recover the reference
ordering of the interface, WT $>$ R248W $>$ R248G in protein--DNA hydrogen
bonds, and both place the mutants above wild type in mean \Ca{} RMSF. Claude
alone additionally reproduces the mutational destabilisation in backbone RMSD;
GPT-5.6 returns R248W as indistinguishable from wild type (1.61 versus
1.62~\AA). Nemotron 3 Ultra inverts the two mutants in hydrogen bonding and
gives its R248W trajectory as the most conformationally stable of all nine runs
(1.52~\AA), the opposite of the expected effect of a destabilising substitution;
its wild-type hydrogen-bond count is also the furthest from the reference
(10.82 against 9.21). The radius of gyration discriminates nothing: all nine
values lie within 0.16~\AA{} of one another, well inside their own standard
deviations. The corresponding framework-level comparison of prompts,
invocations, token usage, wall time, and cost is given in
Table~S6 (Supplementary Information).}

\begin{table*}[htbp]
\centering
\caption{{p53 DBD--DNA observables (mean $\pm$ s.d. over the
100~ns production) from three LLM-orchestrated NAMD3 campaigns, compared with
Balasundaram \& George Priya Doss 2022. All three agents built independently
from PDB 2AC0 with CHARMM36. Every value was recomputed from the raw
trajectories under one identical analysis protocol, so the model columns differ
only in the underlying trajectories.}}
\label{tab:p53_llm}

\normalsize
\renewcommand{\arraystretch}{1.35}
\setlength{\tabcolsep}{9pt}

\begin{adjustbox}{max width=\textwidth}
\begin{tabular}{@{}l l cccc@{}}
\toprule
\textbf{Observable} & \textbf{System}
& \textbf{Claude Opus 4.8} & \textbf{GPT-5.6 Sol}
& \textbf{Nemotron 3 Ultra} & \textbf{Reference paper} \\
\midrule

Backbone RMSD (\AA)
& WT    & $1.77 \pm 0.28$ & $1.62 \pm 0.17$ & $1.80 \pm 0.27$ & $\sim$1.5 \\
& R248W & $1.93 \pm 0.33$ & $1.61 \pm 0.19$ & $1.52 \pm 0.14$ & 1.5--2.5 \\
& R248G & $2.09 \pm 0.66$ & $1.96 \pm 0.23$ & $2.15 \pm 0.38$ & 1.5--2.5 \\
\addlinespace

Mean \Ca{} RMSF (\AA)
& WT    & 0.92 & 0.77 & 0.95 & --- \\
& R248W & 0.98 & 0.83 & 0.91 & $>$ WT \\
& R248G & 1.13 & 0.97 & 1.07 & $>$ WT \\
\addlinespace

Protein--DNA H-bonds
& WT    & $9.15 \pm 1.60$ & $9.37 \pm 2.06$ & $10.82 \pm 1.84$ & 9.21 \\
& R248W & $8.89 \pm 1.93$ & $8.66 \pm 1.72$ & $8.37 \pm 1.66$  & 8.54 \\
& R248G & $8.54 \pm 2.01$ & $8.29 \pm 1.60$ & $8.69 \pm 1.48$  & 6.77 \\
\addlinespace

Protein R$_{g}$ (\AA)
& WT    & $16.83 \pm 0.08$ & $16.88 \pm 0.08$ & $16.78 \pm 0.07$ & 16.94 \\
& R248W & $16.80 \pm 0.10$ & $16.94 \pm 0.08$ & $16.90 \pm 0.08$ & 16.97 \\
& R248G & $16.85 \pm 0.13$ & $16.85 \pm 0.10$ & $16.79 \pm 0.14$ & 16.98 \\

\bottomrule
\end{tabular}
\end{adjustbox}
\end{table*}

\subsection{\textbf{Protein--membrane}: iperoxo stays bound through the canonical M2 orthosteric fingerprint.}

The active-state M2 receptor with iperoxo, embedded in a POPC/cholesterol bilayer (60{,}642 atoms), produced a stable 100-ns membrane-NPT trajectory: the temperature was $303.9 \pm 1.2$~K with negligible energy and volume drift, the \Ca{} RMSD was 2.13~\AA, the radius of gyration was $20.56 \pm 0.08$~\AA, and the helical content was 82.6\%, indicating that the seven-transmembrane bundle was fully preserved. The pharmacologically decisive result is that iperoxo remained bound for the entire run (ligand heavy-atom RMSD $1.28 \pm 0.20$~\AA) through the canonical M2 orthosteric contacts: the Asp103 (D3.32) anchor for the agonist's quaternary-ammonium cation; the aromatic ``tyrosine lid'' formed by Tyr104, Tyr403, Tyr426, and Tyr430; the Trp400 (W6.48) toggle switch; and a persistent Asn404 (N6.52) hydrogen bond with 89\% occupancy. This interaction fingerprint matches the established structural pharmacology of the M2--iperoxo  {complex reported by Kruse et al.~\cite{kruse2013activation}}, validating the agent's membrane build and ligand parametrization. Membrane-specific observables (deuterium order parameters and bilayer-thickness profiles) were outside the current tool set and are not reported. The full set of trajectory-analysis panels supporting these values---protein and ligand RMSD, RMSF, radius of gyration, ligand contacts, and the hydrogen-bond time series---is provided in Figure~S1 (Supplementary Information).

\subsection{\textbf{Protein--ligand}: stable bound poses, but end-point MM-GBSA does not reproduce the TYK2 affinity ranking.}

For the full set of 16 congeneric TYK2 inhibitors, the agent built, simulated
(10~ns each), and ran end-point MM-GBSA on every complex. {Structurally, the compounds in the series are well behaved: all 16 ligands remained stably bound,} with low ligand heavy-atom
RMSD ($\approx$0.3--0.7~\AA), stable protein RMSD ($\approx$1.1--1.7~\AA), constant
radius of gyration ($\approx$19.9~\AA), and a conserved binding fingerprint
(18 pocket residues contacted in $>$50\% of frames by every ligand). The binding poses held throughout.

The MM-GBSA binding energies, however, did \emph{not} reproduce the experimental
affinity ranking across the full series: \textbf{Pearson $r = 0.31$
(Spearman $\rho = 0.29$, $n = 16$)}. An initial five-ligand subset had shown an
apparently strong correlation ($r = 0.81$, $\rho = 0.60$); extending the calculation
to all 16 compounds revealed this to be a small-sample artifact rather than a genuine
predictive signal. By contrast, the gold-standard Schr\"odinger FEP+ reference
reproduces the experimental affinities {accurately done by wang et al.~\cite{wang2015accurate} }(\textbf{$r = 0.93$, $\rho = 0.93$}, mean absolute error 0.37~kcal/mol; per-perturbation $\Delta\Delta G$ $r = 0.79$ across 24 edges).

This is an instructive negative result. The automated pipeline correctly establishes
that the congeneric series shares a stable, native binding mode, a necessary
precondition for relative free-energy methods, but single-trajectory end-point
MM-GBSA is not an adequate quantitative ranking method for this target. Rigorous alchemical free-energy calculations (FEP+) are required. It also underscores that
ranking metrics must be evaluated on the full compound set: a five-compound
correlation can be badly misleading. The per-ligand pose-stability, contact-fingerprint, and MM-GBSA-versus-experiment panels underlying these conclusions are given in Figure~S2 (Supplementary Information).

\subsection{\textbf{Protein--water}: ubiquitin retains its fold and matches the crystallographic radius of gyration.}

{As a protein--water reference, Ubiquitin was benchmarked against its high-resolution crystal structure (PDB 1UBQ) reported by Vijay-Kumar et al.~\cite{vijay1987structure} and against previously reported explicit-solvent MD behavior of 1UBQ~\cite{long2011silico}. The simulation produced a thermodynamically stable trajectory and preserved the characteristic five-stranded $\beta$-sheet and central $\alpha$-helix of the ubiquitin $\beta$-grasp fold.} ubiquitin in explicit water produced a thermodynamically
stable trajectory ($299.2 \pm 2.3$~K; negligible total-energy and volume drift after
equilibration). The fold was fully maintained: the five-stranded $\beta$-sheet and the
central $\alpha$-helix of the $\beta$-grasp fold were retained throughout (per-residue
secondary-structure analysis), and the \Ca{} RMSD stayed in the $\sim$1.2--1.7~\AA{}
range expected for this rigid protein. The most directly verifiable quantity, the
radius of gyration, was $11.98 \pm 0.08$~\AA, in excellent agreement with the
experimental/crystallographic value of $\approx$11.9--12.0~\AA. (We deliberately omit a per-residue RMSF profile here: the available RMSF had not been rotationally aligned and therefore reflected whole-molecule tumbling rather than internal flexibility, an example of the agent flagging an analysis artifact rather than reporting a misleading
number.) The RMSD, radius-of-gyration, SASA, and hydrogen-bond time series supporting this assessment are given in Figure~S3 (Supplementary Information).

\subsection{\textbf{Protein--protein}: barnase--barstar stays natively bound.}

{The barnase--barstar complex, a well-characterized high-affinity protein--protein interaction, reproduced the structural behavior reported in the molecular-dynamics benchmark of Martin and Frezza~\cite{martin2022dynamical}.} Over the production trajectory, the
two partners stayed folded and natively associated: per-chain backbone RMSDs of
1.27~\AA{} (barnase) and 0.85~\AA{} (barstar), matching the benchmark's per-chain
values of $\approx$1~\AA, and an interface RMSD of 1.93~\AA{} (near-native; the
benchmark resolved interface substates at $\sim$0.8 and $\sim$1.5~\AA). The native
hotspot contacts were retained throughout barnase Arg59/Arg83/Arg87/His102/Tyr103
against barstar Asp35/Asp39/Trp44, consistent with the benchmark's conclusion that this
complex remains a stable, native interface on the simulation timescale. The per-chain RMSD, interface RMSD, buried-SASA, and interface-contact panels underlying these numbers are given in Figure~S4 (Supplementary Information).

\vspace{15pt}

Taken together, these results show that the agent reproduced the established behavior of every system class. The full per-system trajectory analyses for the four non-p53 cases
(protein--membrane, protein--ligand, protein--water, and protein--protein) are
collected in the Supplementary Information
(Figures S1-S4), and the natural-language prompts issued to the agent for each case are listed in the Agent prompts portion of the Supplementary Information.

\section{Conclusions}
\label{sec:conclusion}

In this work, we demonstrated that an LLM-driven molecular dynamics agent can move beyond simple input-file generation and support a broader, more self-validating simulation workflow. By building the NAMD-Agent framework, the proposed system can set up, run, analyze, and benchmark simulations across several major classes of biomolecular systems, including protein--DNA, protein--membrane, protein--ligand, solvated protein, and protein--protein systems. Across the validated case studies, the agent reproduced the expected physical and biological trends and clearly identified the case in which the choice of force field affected the outcome. This combination of system breadth, automated benchmark comparison, and self-repair represents an important step from agentic MD workflows that merely generate simulation inputs toward workflows that can deliver checked, interpretable, and reproducible scientific results. At the same time, the results also highlight the continued importance of human oversight, particularly in decisions involving replicate sampling, force-field selection, analysis definitions, and interpretation of incomplete or withheld benchmark artifacts.

The protein--DNA case provides the clearest example of the framework's scientific reliability. By simulating wild-type p53 together with the R248W and R248G hotspot mutants under the same build and simulation protocol, the agent independently recovered the established mechanism of p53 loss of function. Specifically, it captured the progressive weakening of the protein--DNA hydrogen-bond network, with average hydrogen bonds decreasing from 8.24 in the wild type to 7.83 in R248W and 7.15 in R248G. It also identified increasing mobility in the DNA-binding loops in the order WT<R248W<R248G, while confirming that the overall protein fold remained stable with $R_g \approx 16.8$~\AA.

Together, these results correctly rank R248G as the most disruptive variant and show quantitative agreement with the reference study once analysis settings were matched.

Importantly, the agent not only reports numerical agreement but also helps explain the sources of remaining differences. Variations in absolute RMSD values and hydrogen-bond counts were traced to differences in simulation engine, thermostat settings, and hydrogen-bond definitions rather than to a failure of the workflow itself. Similarly, the single anomalous RMSD value observed for the wild-type replica was correctly interpreted as a sampling-related fluctuation from a single replicate rather than as a systematic error. These findings show that the agentic pipeline can reproduce clinically meaningful biomolecular behavior while also providing transparent reasoning about where and why its results may differ from an independent benchmark.

{Running the same protein–DNA study under three different orchestrating models showed that the framework and the model driving it are not the same thing. All three carried the pipeline to completion without human intervention, but they did not arrive at the same science: only Claude Opus-4.8 reproduced every trend of the reference study, while Nemotron 3 Ultra reversed the ordering of the two mutants. What separated them most, though, was not accuracy but cost, which spanned two orders of magnitude for identical simulations and tracked how much context each agent re-sent rather than how much it reasoned. We therefore treat the orchestrating model as part of the method: its identity, parameters, and token accounting belong in the record alongside the force field and the integration timestep.}

Overall, this study supports the use of LLM-guided, tool-connected MD agents as practical assistants for reproducible biomolecular simulation. By combining natural-language interaction with deterministic simulation tools, automated analysis, benchmark comparison, and error-aware reasoning, the framework lowers the technical barrier to MD while preserving the scientific checks needed for trustworthy results. Future work should expand the number of replicates and system types, incorporate additional force fields and simulation engines, and strengthen automated uncertainty quantification so that agentic MD workflows can become more reliable, generalizable, and useful for routine computational biology research.

\subsection*{Limitations}
\begin{enumerate}
 
  \item \textbf{Engine specificity:} While the concepts are engine-agnostic,
        NAMD-Agent presently supports only NAMD/HTMD combinations.
  \item \textbf{LLM hallucination and error-handling:} Although mitigated by
        code retrieval and iterative testing, occasional hallucinated
        parameters (e.g.\ unsupported thermostat keywords) still arise and
        require manual oversight.
  \item \textbf{Scalability:} Large, membrane-embedded systems
        ($>10^{6}$ atoms) stress both browser automation and GPU memory,
        suggesting the need for tighter HPC integration and resource-aware
        planning.
 \item \textbf{User Reliability:} The user is responsible for ensuring that the simulation parameters conform with reality; otherwise, HTMD might terminate the process.
\end{enumerate}

\subsection*{Future Work}
Future work aims to enhance the robustness and applicability of the framework across multiple dimensions. First, expanding multi-engine generalisation by enabling the agent to generate inputs for GROMACS, AMBER, and OpenMM through an independent API would improve versatility greatly. Second, incorporating adaptive simulation steering by coupling online trajectory analysis with reinforcement learning or Bayesian optimisation could allow dynamic control over parameters such as temperature ramps, bias potentials, and sampling windows to better explore rare events. Third, ensuring reproducibility and transparency through end-to-end provenance tracking and packaging the agent within FAIR-compliant workflow descriptions such as CWL or Nextflow, along with containerised deployments, would facilitate reliable use across computing environments. Finally, improving reliability through the integration of multiple specialised language models for tasks including code generation, domain expertise, and error checking, organised under a voting or supervisory framework, could reduce hallucinations and support automatic verification of thermodynamic and topological consistency.

By pursuing these avenues, we anticipate that future incarnations of
NAMD-Agent will evolve from a labour-saving assistant into a \emph{proactive
scientific collaborator}, capable not only of automating established MD
protocols but of designing, executing, and interpreting adaptive simulation
campaigns that push the frontier of computational structural biology.

\section*{Data availability}

The source code is publicly available on GitHub at
\url{https://github.com/BaratiLab/NAMD_AGENT}.
The data underlying this article, including the code files, input files,
and trajectory files, are publicly available in the Zenodo repository, record 22258912, DOI: \href{https://doi.org/10.5281/zenodo.22258912}
{10.5281/zenodo.22258912}. The Zenodo record is available at \url{https://zenodo.org/records/22258912}.

\vspace{15pt}

\section*{Author Contributions} O.B.F. and A.C. performed software development and testing and did benchmarking. A.C. and O.B.F. carried out the result analysis. Writing of the manuscript was done by A.C. and O.B.F.; A.B.F. conceived and approved the manuscript.

\section*{Conflicts of interest}
\label{sec:conflicts}
The authors declare no conflicts of interest.

\section*{\textbf{Acknowledgments}}
\label{sec:acknowledgements}
The authors would like to acknowledge the Mechanical and AI Lab at Carnegie Mellon University for computational resources.

\balance
\bibliographystyle{rsc}
\bibliography{cas-refs}

% End the main article before starting the standalone-style SI.
\label{MainArticleLastPage}
\clearpage
\nobalance
\newgeometry{left=1.8cm,right=1.8cm,top=0.35cm,bottom=1.7cm}
\onecolumn
\hypersetup{hidelinks}
\setstretch{1}
% Restore the SI's 11pt font sizes after the 9pt journal article.
\makeatletter
\renewcommand{\normalsize}{\@setfontsize\normalsize{10.95pt}{13.6pt}}
\renewcommand{\small}{\@setfontsize\small{10pt}{12pt}}
\renewcommand{\footnotesize}{\@setfontsize\footnotesize{9pt}{11pt}}
\renewcommand{\large}{\@setfontsize\large{12pt}{14pt}}
\renewcommand{\Large}{\@setfontsize\Large{14.4pt}{18pt}}
\renewcommand{\LARGE}{\@setfontsize\LARGE{17.28pt}{22pt}}
\makeatother
\normalsize
\sectionfont{\normalfont\fontsize{14.4}{18}\selectfont\bfseries}
\subsectionfont{\normalfont\fontsize{12}{14}\selectfont\bfseries}
\subsubsectionfont{\normalfont\normalsize\bfseries}
\fancypagestyle{plain}{%
  \fancyhf{}%
  \fancyfoot[C]{\thepage}%
  \renewcommand{\headrulewidth}{0pt}%
  \renewcommand{\footrulewidth}{0pt}%
}
% Supporting Information body from SI.tex follows.

\renewcommand*\rmdefault{bch}\normalfont\upshape
\setcounter{secnumdepth}{1}
\setcounter{tocdepth}{1}
\setcounter{page}{1}
\renewcommand{\thepage}{S\arabic{page}}
\setcounter{section}{0}
\setcounter{table}{0}
\setcounter{figure}{0}
\renewcommand{\thesection}{S\arabic{section}}
\renewcommand{\thetable}{S\arabic{table}}
\renewcommand{\thefigure}{S\arabic{figure}}
\renewcommand{\theHsection}{supp.section.\arabic{section}}
\renewcommand{\theHtable}{supp.table.\arabic{table}}
\renewcommand{\theHfigure}{supp.figure.\arabic{figure}}
\pagestyle{plain}
\thispagestyle{plain}

\noindent{\sffamily\fontsize{10.5}{11.5}\selectfont
Supplementary Information (SI) for Digital Discovery.\\
This journal is \textcopyright\ The Royal Society of Chemistry 2026}

\vspace*{2.8cm}
\begin{center}
{\sffamily\bfseries\fontsize{18}{22}\selectfont Supporting Information:\par}
\vspace{0.65cm}
{\sffamily\bfseries\fontsize{18}{24}\selectfont
Automating MD simulations for Proteins using Large language Models:\par
Benchmark-Validated NAMD-Agent\par}

\vspace{0.8cm}
{\sffamily\fontsize{12}{19}\selectfont
Omid Barati Farimani,$^{a}$ Achuth Chandrasekhar,$^{a}$ and\\
Amir Barati Farimani$^{\ast,a,b,c,d}$\par}

\vspace{0.65cm}
{\itshape\fontsize{10.5}{17}\selectfont
$^{a}$Department of Mechanical Engineering, Carnegie Mellon University,
Pittsburgh, PA 15213, USA.\\
$^{b}$Department of Biomedical Engineering, Carnegie Mellon University,
Pittsburgh, PA 15213, USA.\\
$^{c}$Department of Chemical Engineering, Carnegie Mellon University,
Pittsburgh, PA 15213, USA.\\
$^{d}$Machine Learning Department, Carnegie Mellon University,
Pittsburgh, PA 15213, USA.\par}

\vspace{0.55cm}
{\sffamily\fontsize{10.5}{13}\selectfont
$^{\ast}$E-mail: barati@cmu.edu\par}
\end{center}
\vspace{0.65cm}
{\Large\bfseries Contents\par}
\vspace{0.55cm}

\noindent
\begin{tabularx}{\textwidth}{@{}Xr@{}}
S1\quad \texttt{Five benchmark source data} & \pageref{si:source} \\[1.15em]
S2\quad \texttt{htmd-mcp} tool catalog & \pageref{si:tools} \\[1.15em]
S3\quad Per-system trajectory analysis (non-p53 cases) & \pageref{si:analysis} \\[1.15em]
S4\quad Agent prompts & \pageref{si:prompts} \\[1.15em]
S5\quad Measured setup cost and error ledger & \pageref{si:setupcost} \\[1.15em]
S6\quad Cross-model orchestration comparison (protein--DNA) & \pageref{si:llm} \\
\end{tabularx}

\clearpage

\section{\texttt{Five benchmark source data}}
\label{si:source}

{The source data for the five benchmark systems were extracted from their corresponding reference studies and are summarized in Table~\ref{tab:benchmark_system_preparation}. The table accounts for every modification between the deposited PDB entry and the final system that entered the energy-minimization stage.}

\begin{table*}[htbp]
\centering
\caption{{Summary of the structural preprocessing and system-building steps for the five benchmark systems. The table accounts for all modifications between the deposited PDB structures and the final systems used for energy minimization.}}
\label{tab:benchmark_system_preparation}

\footnotesize
\setlength{\tabcolsep}{3pt}
\renewcommand{\arraystretch}{1.25}

\begin{tabularx}{\textwidth}{
    >{\raggedright\arraybackslash}p{2.1cm}
    >{\raggedright\arraybackslash}p{1.8cm}
    >{\raggedright\arraybackslash}X
    >{\raggedright\arraybackslash}p{2.2cm}
    >{\raggedright\arraybackslash}X
    >{\raggedright\arraybackslash}p{1.8cm}
    >{\raggedright\arraybackslash}X
}
\hline
\textbf{System (PDB)} &
\textbf{Deposited} &
\textbf{Chains kept} &
\textbf{Crystallographic waters removed} &
\textbf{Hetero records removed} &
\textbf{Alternate-location atoms discarded} &
\textbf{Built system} \\
\hline

Ubiquitin (1UBQ) &
660 atoms, 1 chain &
A &
58 water atoms (all) &
None &
0 &
17,215 atoms; 5,315 waters, 15 Na$^{+}$, 15 Cl$^{-}$ \\

Barnase--barstar (1BRS) &
5,153 atoms, 6 chains &
C (barnase) + F (barstar); 4 redundant copies removed &
513 water atoms (all) &
None &
2 of 4 &
29,580 atoms; 8,792 waters, 29 K$^{+}$, 25 Cl$^{-}$ \\

RNase Sa--barstar (1AY7) &
1,678 atoms, 2 chains &
A + B &
190 water atoms (all) &
None &
22 of 44 &
24,956 atoms; 7,336 waters, 34 K$^{+}$, 21 Cl$^{-}$ \\

p53 DBD--DNA (2AC0) &
8,247 atoms, 8 chains &
A (protein) + E, F (DNA duplex); 5 chains removed &
1,054 water atoms (all) &
3 of 4 Zn$^{2+}$ removed; 1 structural Zn$^{2+}$ retained &
37 of 74 &
47,090 atoms; 14,383 waters, 57 Na$^{+}$, 41 Cl$^{-}$, 1 Zn$^{2+}$ \\

M2--iperoxo (4MQS) &
3,034 atoms, 2 chains &
A (receptor) + IXO ligand; chain B (nanobody Nb9-8, 892 atoms) removed &
None present in entry &
None &
0 &
60,642 atoms; 130 POPC, 17 cholesterol, 12,450 waters, 35 Na$^{+}$, 46 Cl$^{-}$ \\

\hline
\end{tabularx}
\end{table*}

The p53 protein--DNA analyses are presented in the main paper. This section collects (i) the \texttt{htmd-mcp}
tool catalog, (ii) the per-system trajectory analyses for the four remaining
cases, and (iii) the natural-language prompts issued to the agent. Unless
stated otherwise, systems were built with CHARMM36m (protein), CHARMM36
(lipid/nucleic), and CGenFF/SwissParam (ligands) in TIP3P water, neutralized to
0.15~M salt, and run with NAMD3.

\section{htmd-mcp tool catalog} 
\label{si:tools}

The server registers 45 tools spanning the full simulation lifecycle.
Table~\ref{tab:tools} lists the system-construction tools in pipeline order, and
Table~\ref{tab:analysis} lists the trajectory-analysis and comparison tools.

\begin{table}[htbp]
\centering
\caption{System-construction tools in the \texttt{htmd-mcp} server and their tasks, in
pipeline order. Steps chain automatically: each builder emits a matched PSF/PDB, and
each setup tool stages parameters and inherits the previous stage's coordinates,
velocities, and box.}
\label{tab:tools}
\small
\begin{tabular}{@{}p{2.4cm} p{3.6cm} p{8.4cm}@{}}
\toprule
Stage & Tool & Task \\
\midrule
Acquire\&inspect
 & \texttt{fetch\_pdb} & Resolve a PDB ID or path to a local file with a chain summary \\
 & \texttt{inspect\_structure} & Audit chains, segids, ligands, ions, waters, and chain gaps \\
 & {\fontsize{9pt}{10pt}\selectfont \texttt{fetch\_ligand\_from\_rcsb}} & Retrieve a ligand by 3-letter code as SDF/mol2 \\
\addlinespace
Prepare \& clean
 & \texttt{prepare\_protein} & Protonation (PROPKA), H-bond optimization, missing-atom rebuild \\
 & \texttt{clean\_structure} & Remove or keep atoms by atomselect (strip buffers, reference ligands) \\
 & \texttt{auto\_segment} & Assign segids for \texttt{psfgen} \\
\addlinespace
Ligand     \- parameters
 & \texttt{prepare\_ligand\_mol2} & Standardize any ligand input to a parameter-ready mol2 (explicit H, 3D, charges) \\
 & \texttt{parametrize\_ligand} & CGenFF/SwissParam CHARMM topology and parameters \\
\addlinespace
Build solvated system
 & {\fontsize{9pt}{10pt}\selectfont\texttt{build\_protein\_waterbox}} & Soluble protein in a water box \\
 & \texttt{build\_protein\_ligand} & Protein + ligand (docked or unbinding pose), solvated \\
 & {\fontsize{8pt}{10pt}\selectfont\texttt{build\_protein\_multiligand}} & Protein + multiple small molecules \\
 & {\fontsize{8pt}{10pt}\selectfont\texttt{build\_protein\_membrane}} & Protein embedded in a lipid bilayer (with \texttt{list\_lipids}) \\
 & {\fontsize{9pt}{10pt}\selectfont \texttt{build\_protein\_nucleic}} & Protein + DNA/RNA complex \\
 & {\fontsize{7pt}{8pt}\selectfont \texttt{build\_protein\_protein\_complex}} & Protein--protein complex \\
\addlinespace
Generate run inputs
 & \texttt{setup\_minimization} & Restrained minimization input (\texttt{min.namd}) \\
 & \texttt{setup\_equilibration} & Restraint-ramped equilibration, chained from minimization \\
 & \texttt{setup\_production} & Production input, chained from equilibration (optional HMR for 4~fs) \\
 & \texttt{prepare\_run} & One-shot minimize $\rightarrow$ equilibrate $\rightarrow$ produce \\
\bottomrule
\end{tabular}
\end{table}

\begin{table}[htbp]
\centering
\caption{Trajectory-analysis and comparison tools in the \texttt{htmd-mcp} server. Each
writes CSV data plus a plot into \texttt{run\_dir/analysis/}; the class-specific and
landscape tools are applied only where relevant.}
\label{tab:analysis}
\small
\begin{tabular}{@{}p{2.6cm} p{3.7cm} p{8.1cm}@{}}
\toprule
Category & Tool & Task \\
\midrule
Preprocess
 & \texttt{wrap\_and\_align} & Re-image waters/ions and align the trajectory (removes PBC artifacts) \\
\addlinespace
Global stability
 & \texttt{compute\_rmsd} & Per-frame \Ca{} RMSD versus a reference frame \\
 & \texttt{compute\_rmsf} & Per-residue fluctuation over the trajectory \\
 & \texttt{radius\_of\_gyration} & Per-frame mass-weighted $R_g$ \\
 & {\fontsize{8pt}{8pt}\selectfont\texttt{compute\_secondary\_structure}} & DSSP assignment per residue per frame \\
\addlinespace
Solvation \& contacts
 & \texttt{compute\_sasa} & Solvent-accessible surface area (total or per-residue) \\
 & \texttt{compute\_distances} & Distance between two selections (min / centroid / all-pairs) \\
 & \texttt{analyze\_hbonds} & Hydrogen-bond detection, occupancies, and count time series \\
\addlinespace
Protein--ligand
 & \texttt{ligand\_rmsd} & Ligand heavy-atom RMSD after protein alignment (pose drift) \\
 &{\fontsize{9pt}{10pt}\selectfont \texttt{ligand\_protein\_contacts}} & Per-residue ligand contact frequency (binding fingerprint) \\
 & \texttt{mm\_pbsa} & Single-trajectory MM-GBSA binding free-energy estimate \\
\addlinespace
Interfaces
 & \texttt{interface\_rmsd} & CAPRI-style interface backbone RMSD \\
 & \texttt{interface\_contacts} & Per-residue contact frequency at a protein--protein interface \\
 & \texttt{buried\_sasa} & Interfacial buried SASA (association/dissociation) \\
\addlinespace
Conformational landscape
 & \texttt{compute\_pca} & Cartesian PCA of aligned coordinates \\
 & \texttt{cluster\_trajectory} & Cluster frames in PC or RMSD space \\
 &{\fontsize{9pt}{10pt}\selectfont\texttt{free\_energy\_landscape}} & $-k_{\mathrm B}T\ln P$ over collective variables \\
 & {\fontsize{7pt}{8pt}\selectfont{\texttt{kernel\_density\_estimation}}} & Probability density over collective variables \\
\addlinespace
QA \& visualization
 & {\fontsize{9pt}{10pt}\selectfont\texttt{quality\_assurance\_report}} & Energy/temperature/pressure/volume drift dashboard \\
 & \texttt{bond\_length\_sanity} & Flag broken or stretched bonds in a built system \\
 & {\fontsize{9.5pt}{10pt}\selectfont\texttt{render\_molecular\_movie}} & PyMOL/ffmpeg trajectory movie \\
 & {\fontsize{9.5pt}{10pt}\selectfont\texttt{render\_dashboard\_movie}} & Animated multi-panel timeseries movie \\
\addlinespace
Comparison
 & {\fontsize{9pt}{10pt}\selectfont\texttt{compare\_protein\_nucleic}} & Overlay RMSD/RMSF/$R_g$/H-bonds across systems, mutants, or replicas \\
\bottomrule
\end{tabular}
\end{table}

\FloatBarrier

\section{Per-system trajectory analysis (non-p53 cases)}
\label{si:analysis}

The panels below summarize the production-trajectory analyses for the four
systems whose figures are not shown in the main text. All runs are stable and
reproduce the expected reference behavior described in the Results.

% ---------- Figure S1: M2 membrane ----------
\begin{figure}[htbp]
\centering
\begin{subfigure}[b]{0.32\linewidth}\includegraphics[width=\linewidth]{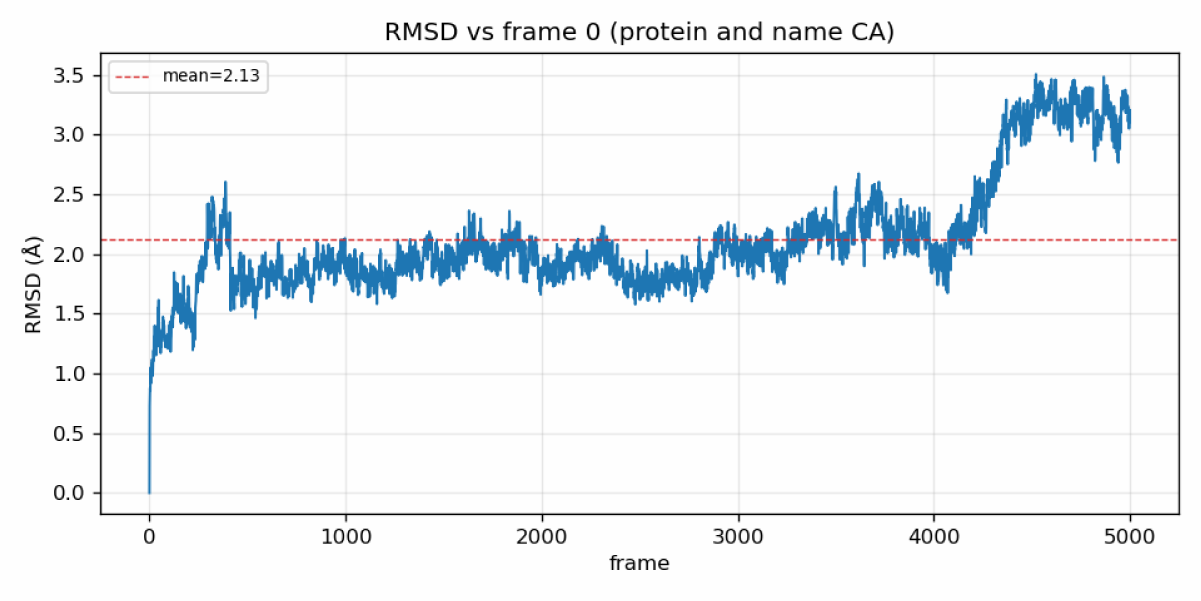}\caption{Protein \Ca{} RMSD}\end{subfigure}\hfill
\begin{subfigure}[b]{0.32\linewidth}\includegraphics[width=\linewidth]{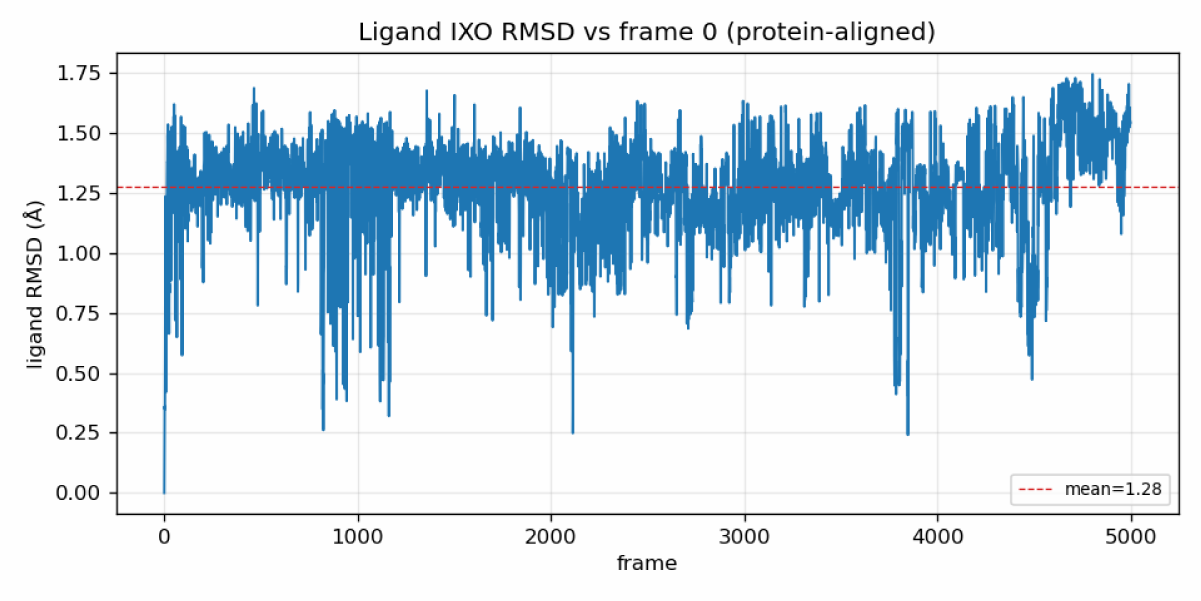}\caption{Iperoxo heavy-atom RMSD}\end{subfigure}\hfill
\begin{subfigure}[b]{0.32\linewidth}\includegraphics[width=\linewidth]{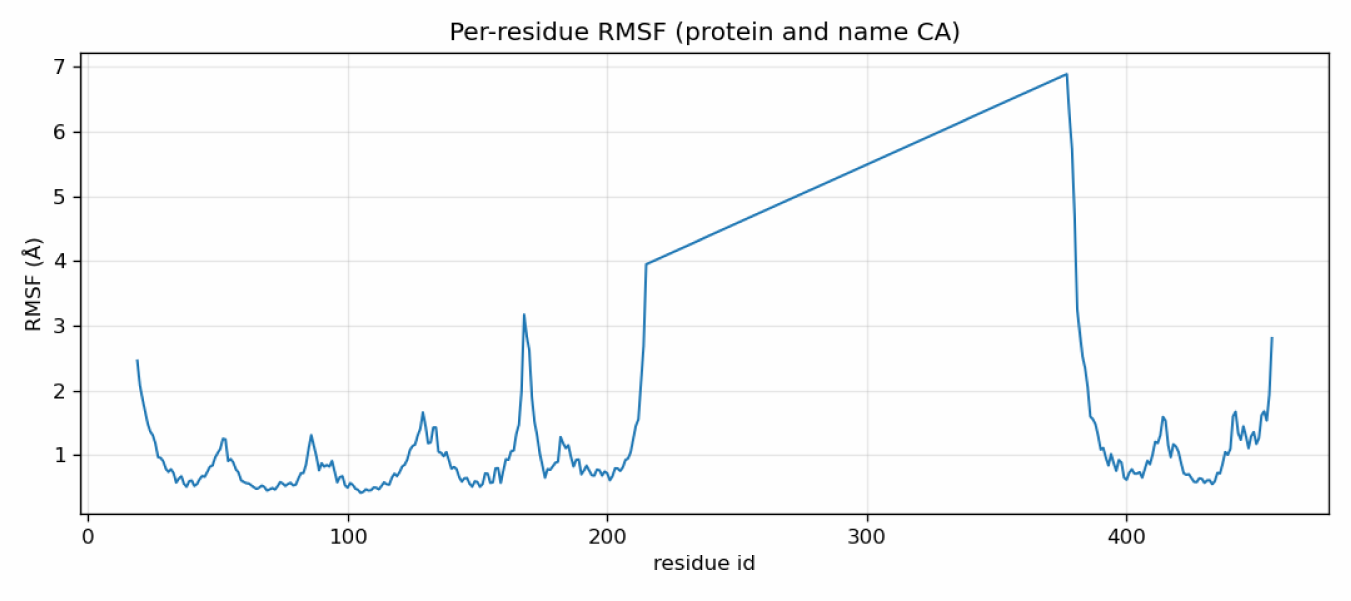}\caption{Per-residue RMSF}\end{subfigure}

\vspace{0.4cm}
\begin{subfigure}[b]{0.32\linewidth}\includegraphics[width=\linewidth]{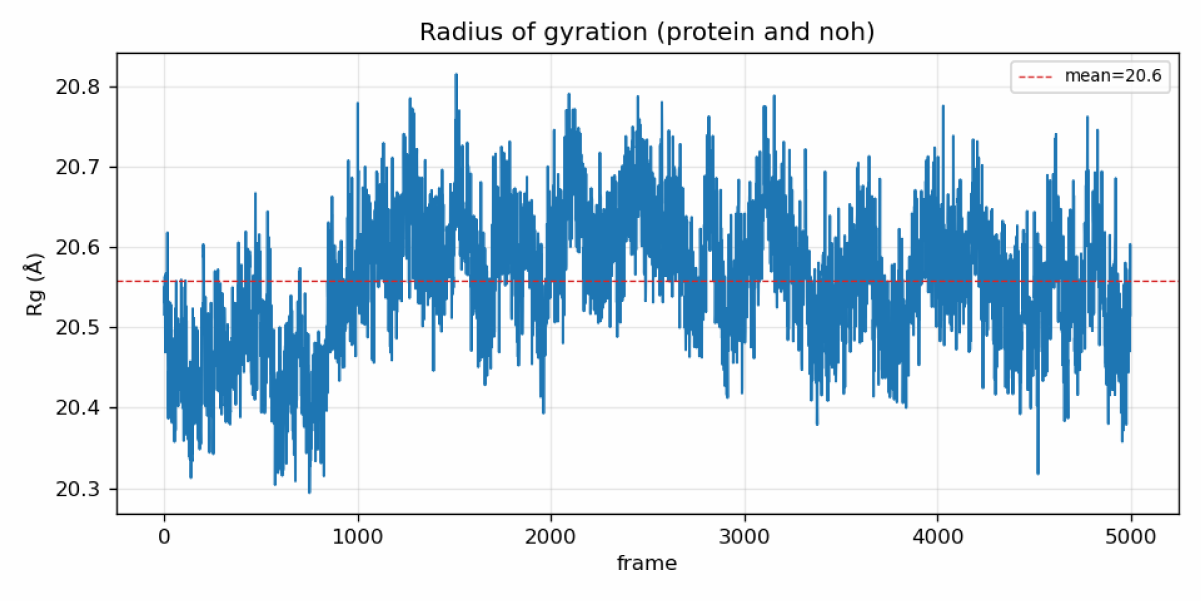}\caption{Radius of gyration}\end{subfigure}\hfill
\begin{subfigure}[b]{0.32\linewidth}\includegraphics[width=\linewidth]{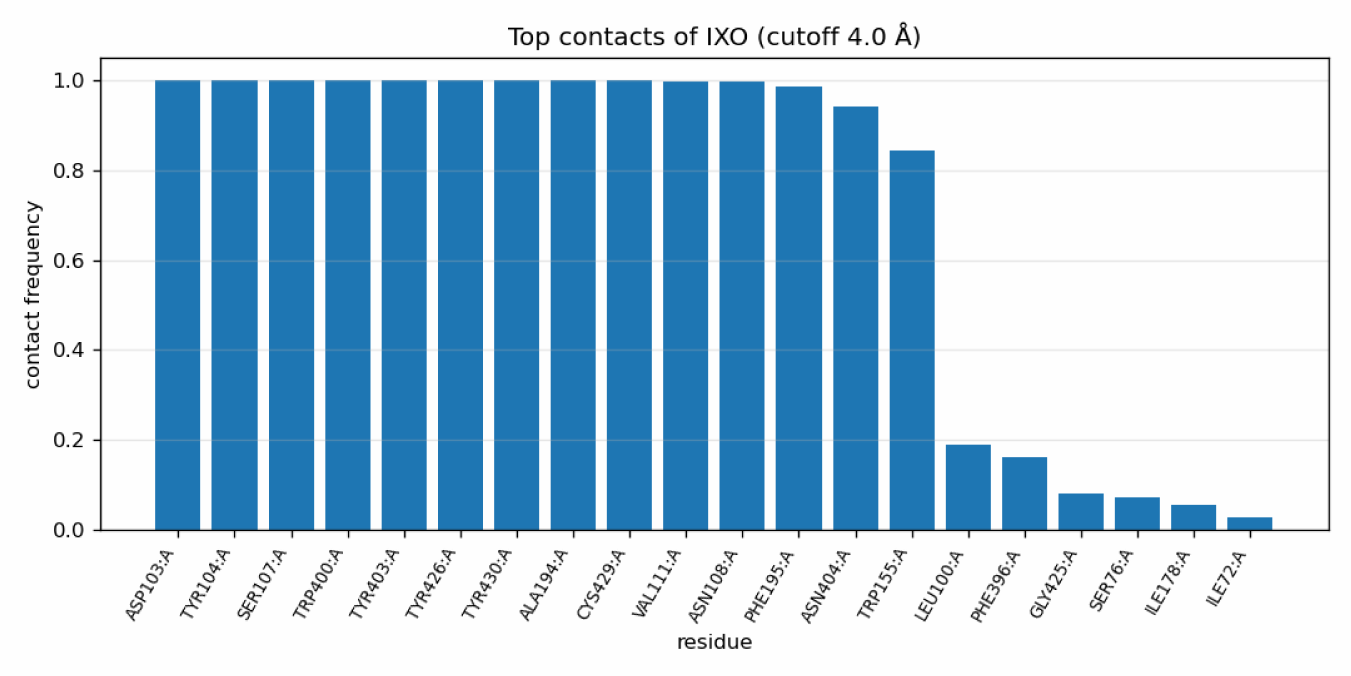}\caption{Ligand--protein contacts}\end{subfigure}\hfill
\begin{subfigure}[b]{0.32\linewidth}\includegraphics[width=\linewidth]{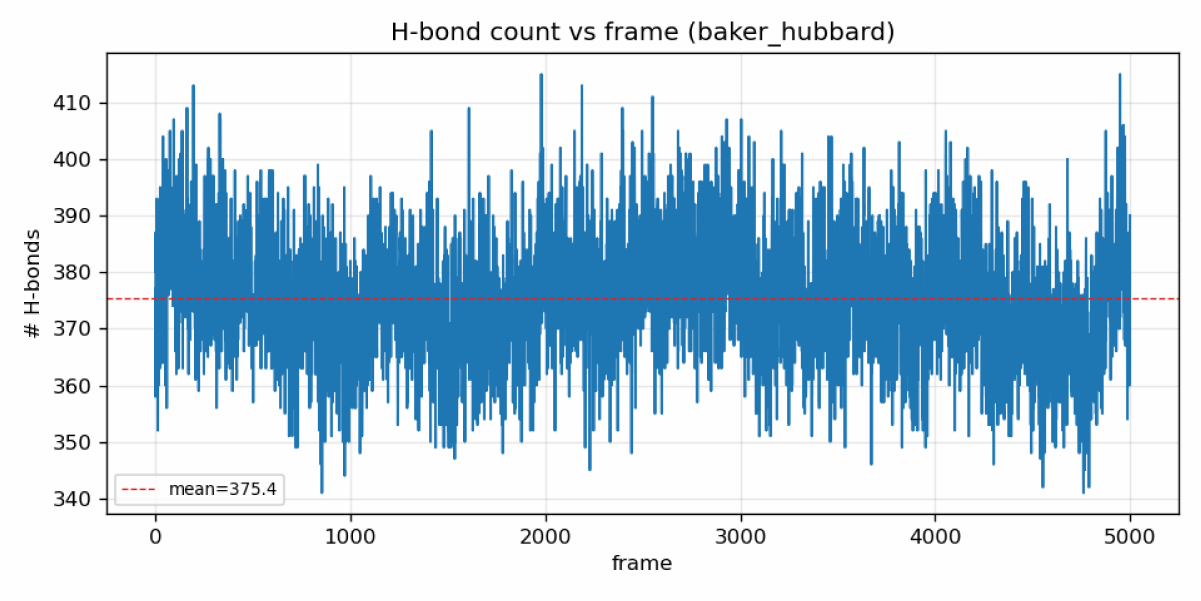}\caption{Protein--ligand H-bonds}\end{subfigure}
\caption{\footnotesize \textbf{Protein--membrane (M2 muscarinic receptor + iperoxo, PDB 4MQS in a POPC/cholesterol 9:1 bilayer, 100~ns).} The 7-TM fold is preserved (\Ca{} RMSD $\sim$2.1~\AA, $R_g$ stable), and iperoxo remains bound (heavy-atom RMSD $\sim$1.3~\AA) through its canonical orthosteric contacts and the Asn404 hydrogen bond.}
\label{fig:si_m2}
\end{figure}

Fig.~S1a shows the protein \Ca{} RMSD, fit to the equilibrated frame, which plateaus at $\sim$2.1~\AA{} after the first $\sim$10~ns, indicating the seven-transmembrane bundle does not unfold over the run. Fig.~S1b gives the iperoxo heavy-atom RMSD after protein alignment, which stays at $1.28\pm0.20$~\AA{} throughout, confirming the agonist never leaves the orthosteric pocket. Fig.~S1c plots the per-residue RMSF, which is low across the transmembrane helices and rises only at the extracellular loops and termini, consistent with a rigid helical bundle whose flexible loops sample a wider ensemble. Fig.~S1d shows the radius of gyration, flat at $20.56\pm0.08$~\AA, indicating the bundle neither swells nor collapses. Fig.~S1e gives the ligand--protein contact map, which recovers the textbook orthosteric fingerprint (Asp103/D3.32, the Y104/Y403/Y426/Y430 tyrosine lid, and Trp400/W6.48). Fig.~S1f shows the Asn404 (N6.52) hydrogen-bond occupancy, which remains high (89\%) throughout, the single most diagnostic contact for an agonist-bound M2 receptor.

% ---------- Figure S2: TYK2 protein-ligand ----------
\begin{figure}[htbp]
\centering
\begin{subfigure}[b]{0.45\linewidth}\includegraphics[width=\linewidth]{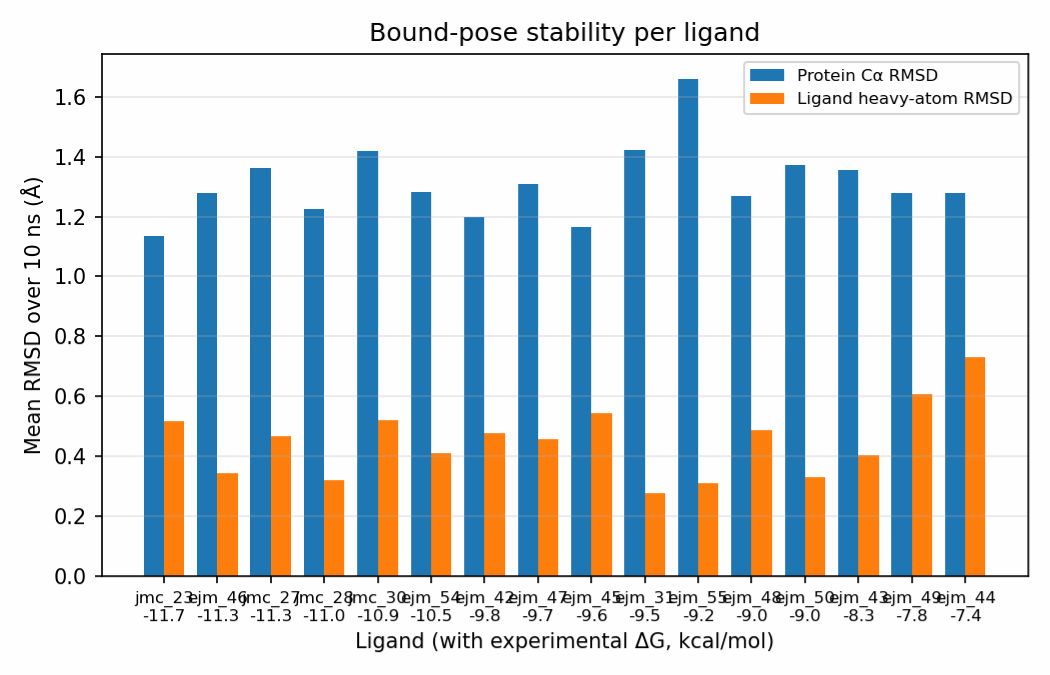}\caption{{Per-ligand pose stability}}\end{subfigure}\hfill
\begin{subfigure}[b]{0.45\linewidth}\includegraphics[width=\linewidth]{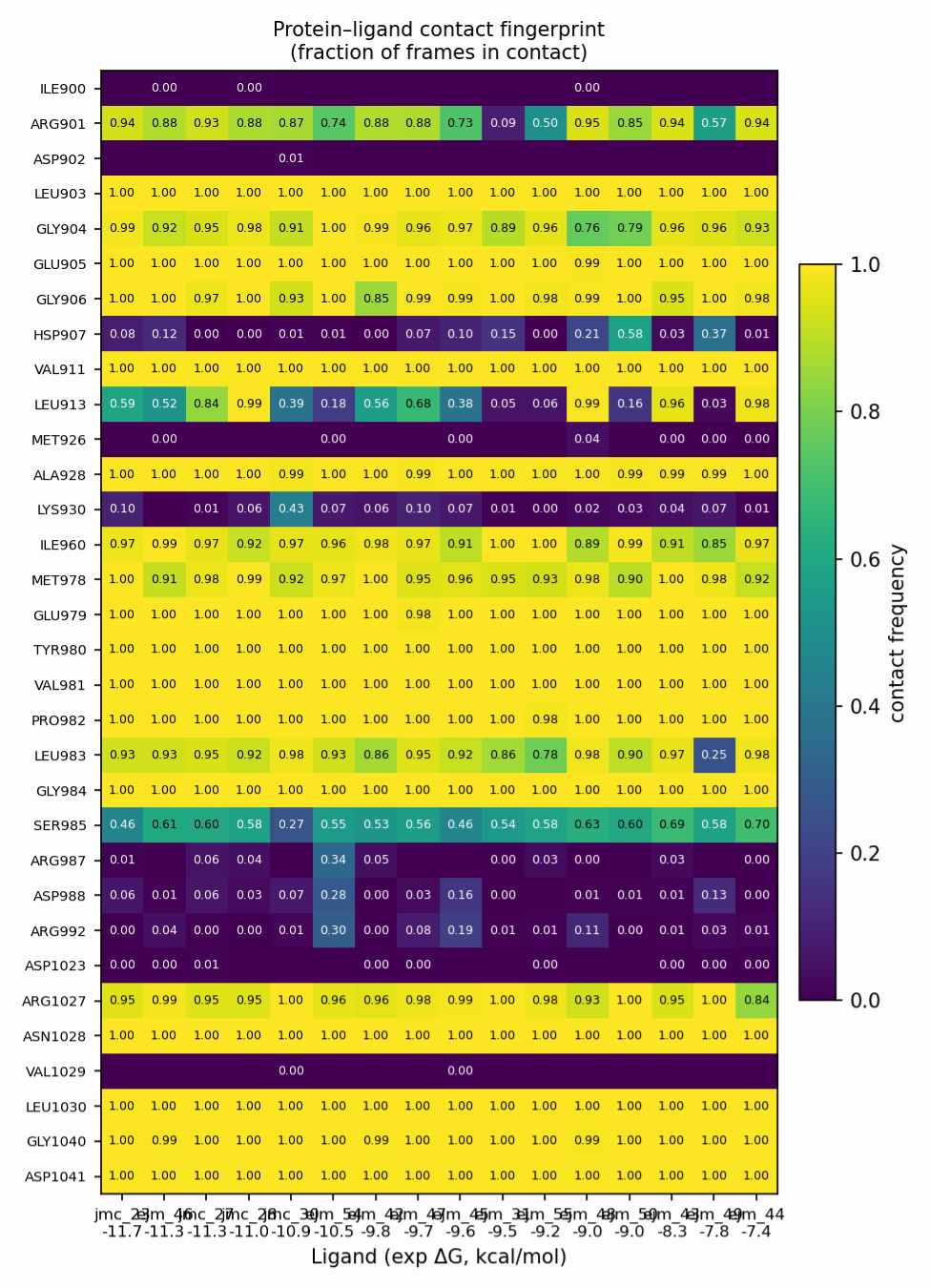}\caption{{Ligand contact fingerprint}}\end{subfigure}

\vspace{0.4cm}
\begin{subfigure}[b]{0.45\linewidth}\includegraphics[width=\linewidth]{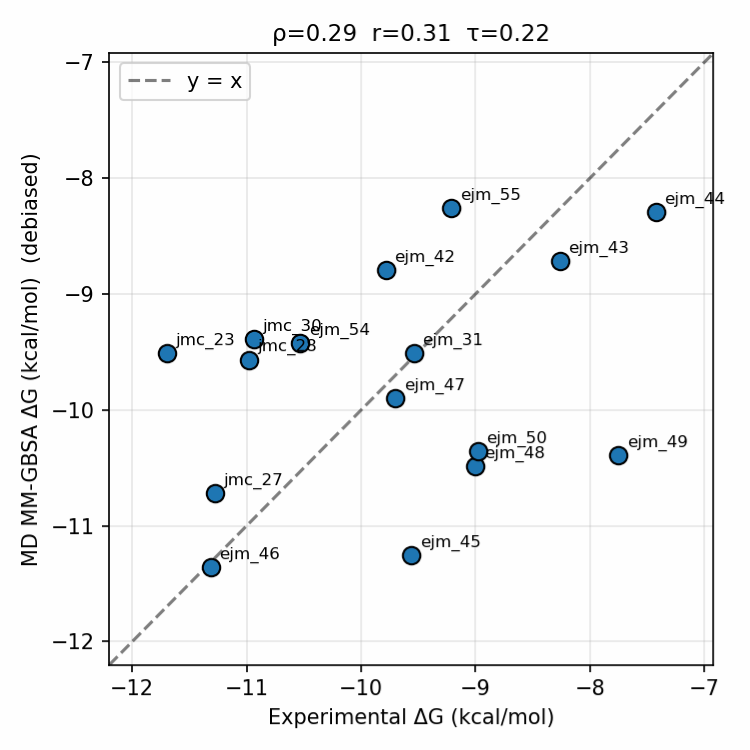}\caption{MM-GBSA vs.\ experiment}\end{subfigure}\hfill
\begin{subfigure}[b]{0.45\linewidth}\includegraphics[width=\linewidth]{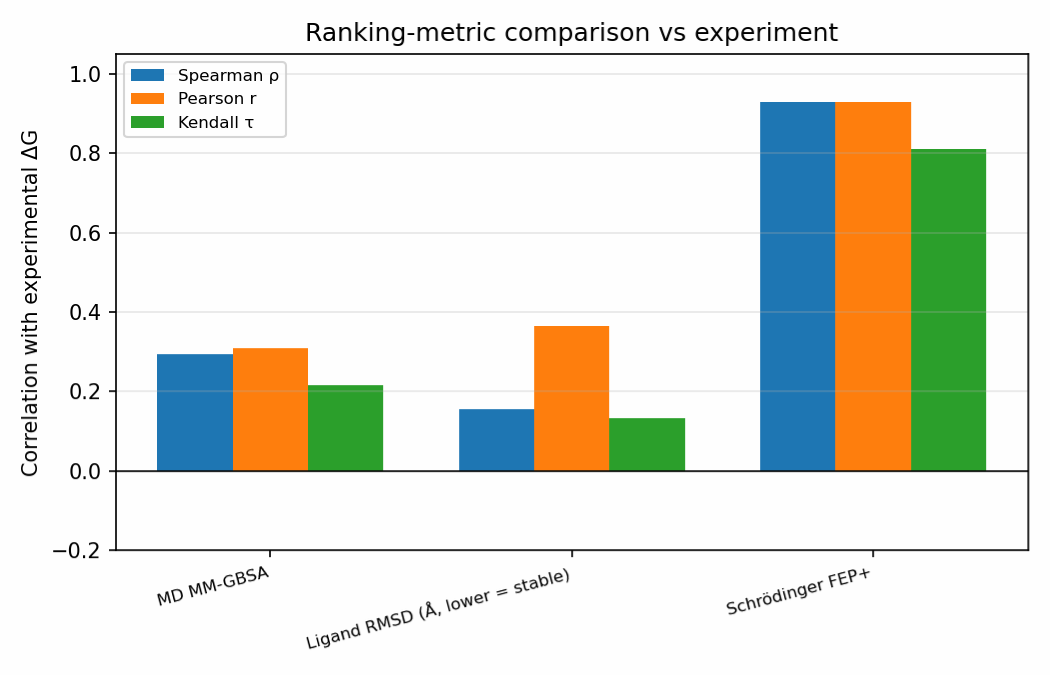}\caption{Ranking correlations}\end{subfigure}
\caption{\footnotesize \textbf{Protein--ligand (16 congeneric TYK2 inhibitors, 10~ns each).} All ligands remain stably bound with a conserved contact fingerprint, but single-trajectory end-point MM-GBSA does not reproduce the experimental affinity ranking ($r=0.31$), unlike the Schr\"odinger FEP+ reference ($r=0.93$).}
\label{fig:si_tyk2}
\end{figure}

Fig.~S2a shows the per-ligand heavy-atom RMSD (pose stability), which stays within $\approx$0.3--0.7~\AA{} for every compound, showing that each inhibitor holds its docked pose for the entire production run. Fig.~S2b gives the per-residue contact fingerprint, essentially identical across the series (18 pocket residues contacted in $>$50\% of frames by every ligand), confirming a shared binding mode across the congeneric set. Fig.~S2c plots the single-trajectory MM-GBSA binding energies against the experimental affinities, showing wide scatter about the diagonal with no consistent linear trend across the full 16-compound series. Fig.~S2d shows Pearson/Spearman/Kendall correlations computed as ligands are added to the set, illustrating how an apparently strong 5-ligand correlation ($r=0.81$, $\rho=0.60$) collapses to $r=0.31$ ($\rho=0.29$) once all 16 compounds are included, in contrast to the FEP+ reference ($r=0.93$).

% ---------- Figure S3: ubiquitin protein-water ----------
\begin{figure}[htbp]
\centering
\begin{subfigure}[b]{0.45\linewidth}\includegraphics[width=\linewidth]{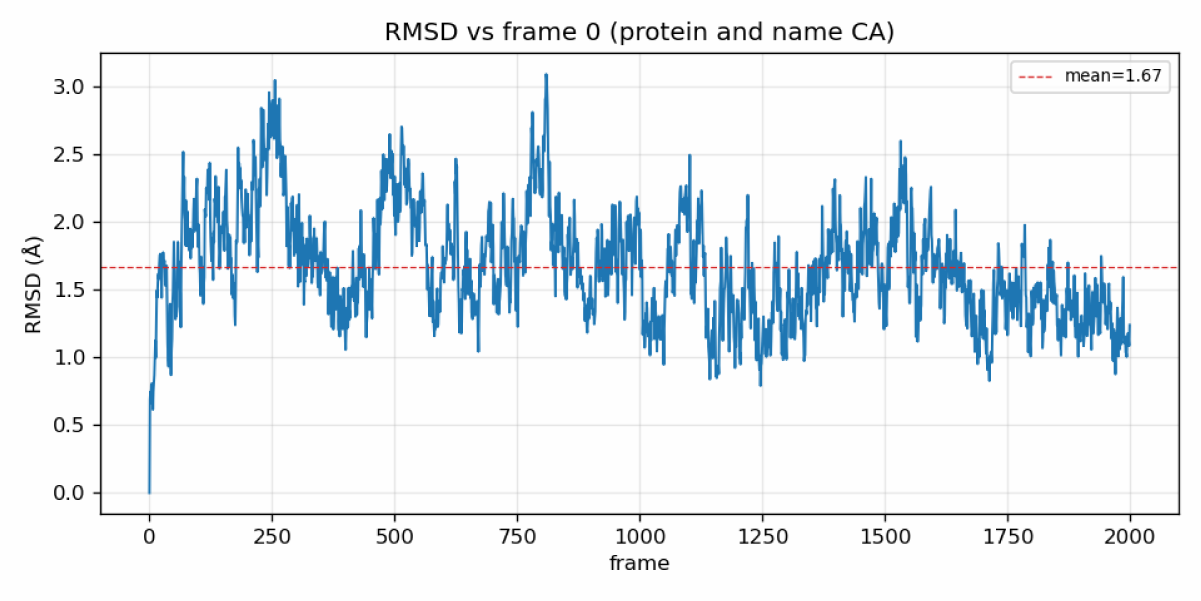}\caption{\Ca{} RMSD}\end{subfigure}\hfill
\begin{subfigure}[b]{0.45\linewidth}\includegraphics[width=\linewidth]{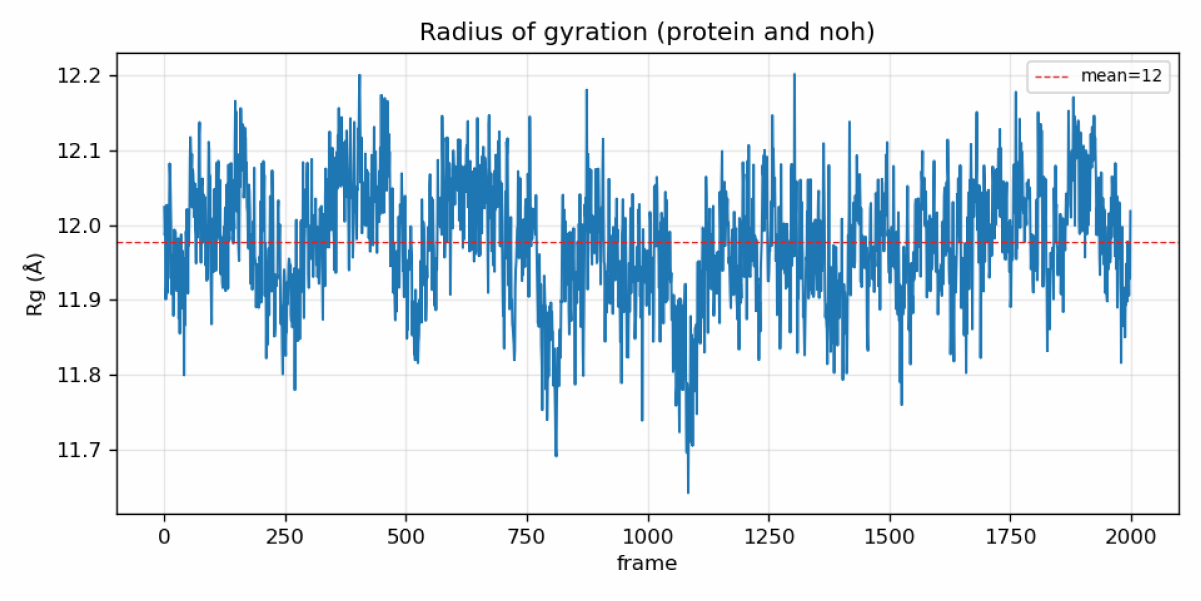}\caption{Radius of gyration}\end{subfigure}

\vspace{0.4cm}
\begin{subfigure}[b]{0.45\linewidth}\includegraphics[width=\linewidth]{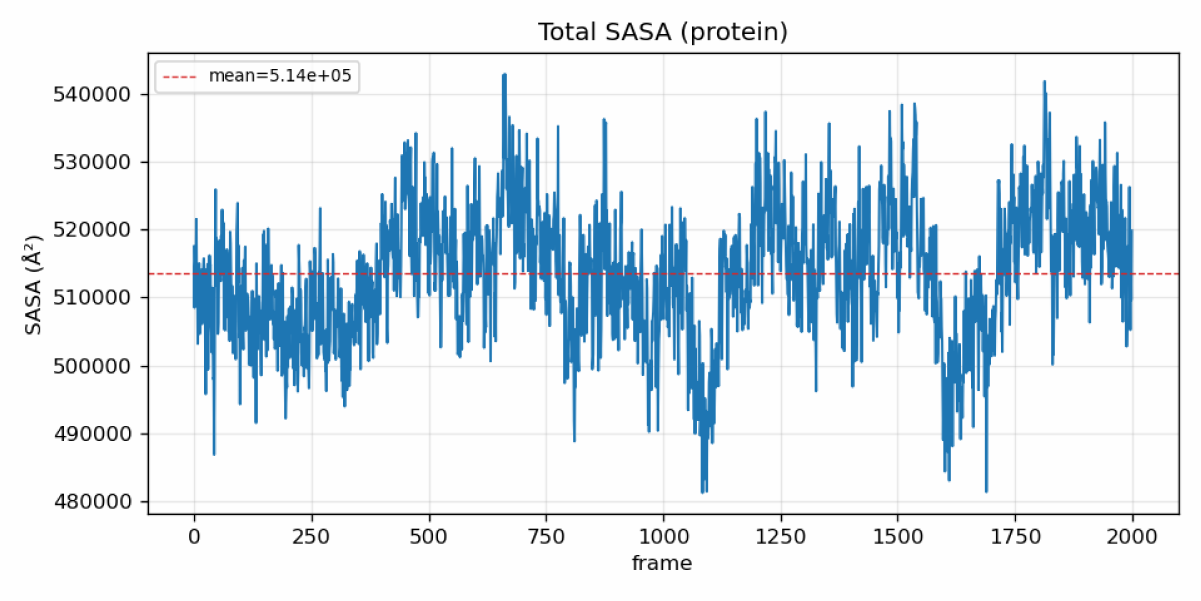}\caption{Solvent-accessible surface area}\end{subfigure}\hfill
\begin{subfigure}[b]{0.45\linewidth}\includegraphics[width=\linewidth]{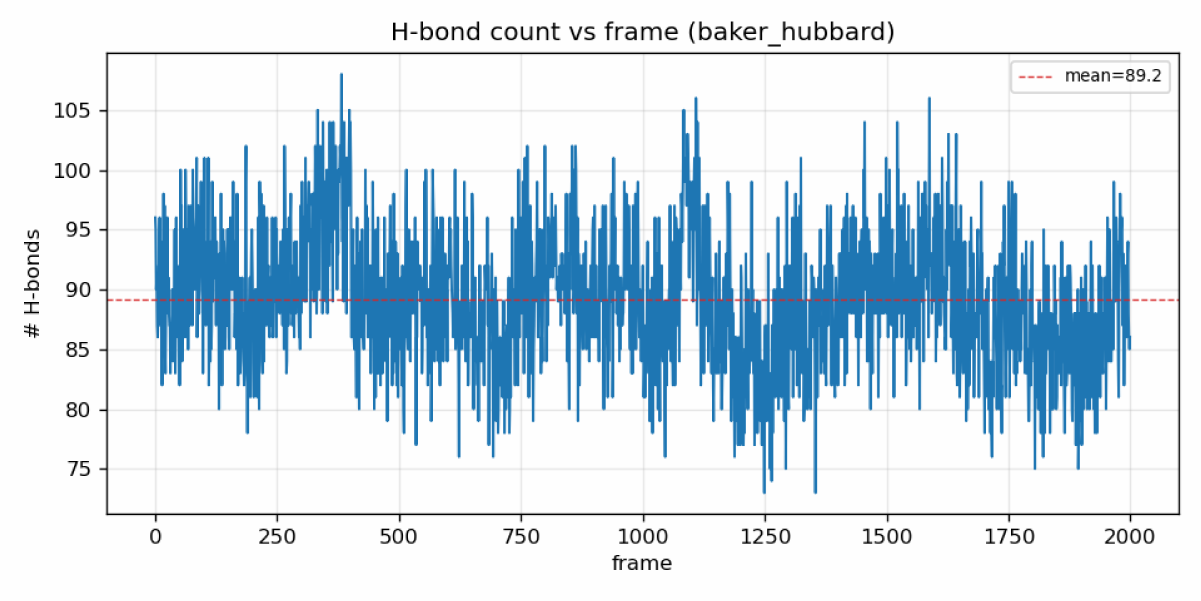}\caption{Hydrogen-bond count}\end{subfigure}
\caption{\footnotesize \textbf{Protein--water (ubiquitin, PDB 1UBQ).} The $\beta$-grasp fold is fully retained; \Ca{} RMSD stays in the $\sim$1.2--1.7~\AA{} range and the radius of gyration ($11.98\pm0.08$~\AA) matches the crystallographic value. Per-residue RMSF is deliberately omitted (the available profile was not rotationally aligned).}
\label{fig:si_ubq}
\end{figure}

Fig.~S3a shows the \Ca{} RMSD, which stays within the $\sim$1.2--1.7~\AA{} range expected for this small, rigid protein, with no drift toward larger values over the trajectory. Fig.~S3b gives the radius of gyration, stable at $11.98\pm0.08$~\AA, matching the experimental/crystallographic value of $\approx$11.9--12.0~\AA{} to within uncertainty. Fig.~S3c plots the total solvent-accessible surface area, which fluctuates about a constant mean, indicating the fold is not partially unfolding or exposing additional hydrophobic core to solvent. Fig.~S3d shows the hydrogen-bond count, including the contacts stabilizing the central $\alpha$-helix and the five-stranded $\beta$-sheet, which remains constant across the run, consistent with retention of the $\beta$-grasp fold.

% ---------- Figure S4: barnase-barstar protein-protein ----------
\begin{figure}[htbp]
\centering
\begin{subfigure}[b]{0.45\linewidth}\includegraphics[width=\linewidth]{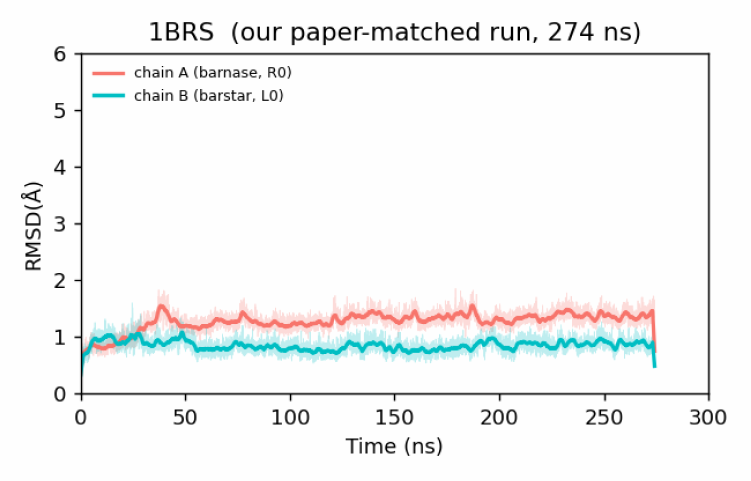}\caption{Per-chain backbone RMSD}\end{subfigure}\hfill
\begin{subfigure}[b]{0.45\linewidth}\includegraphics[width=\linewidth]{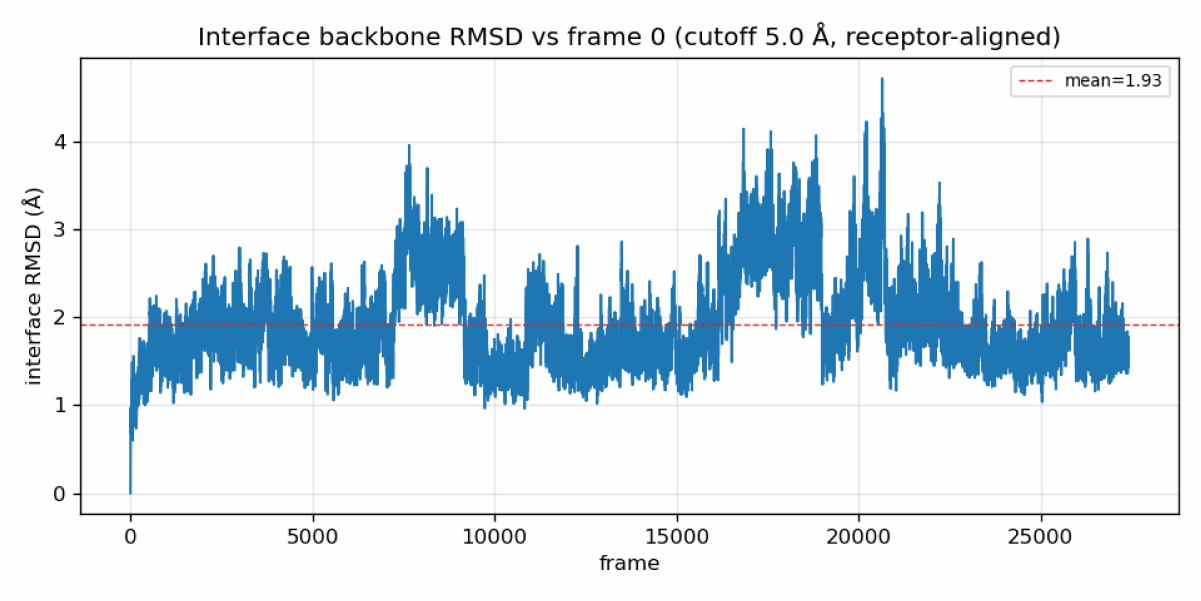}\caption{Interface RMSD}\end{subfigure}

\vspace{0.4cm}
\begin{subfigure}[b]{0.45\linewidth}\includegraphics[width=\linewidth]{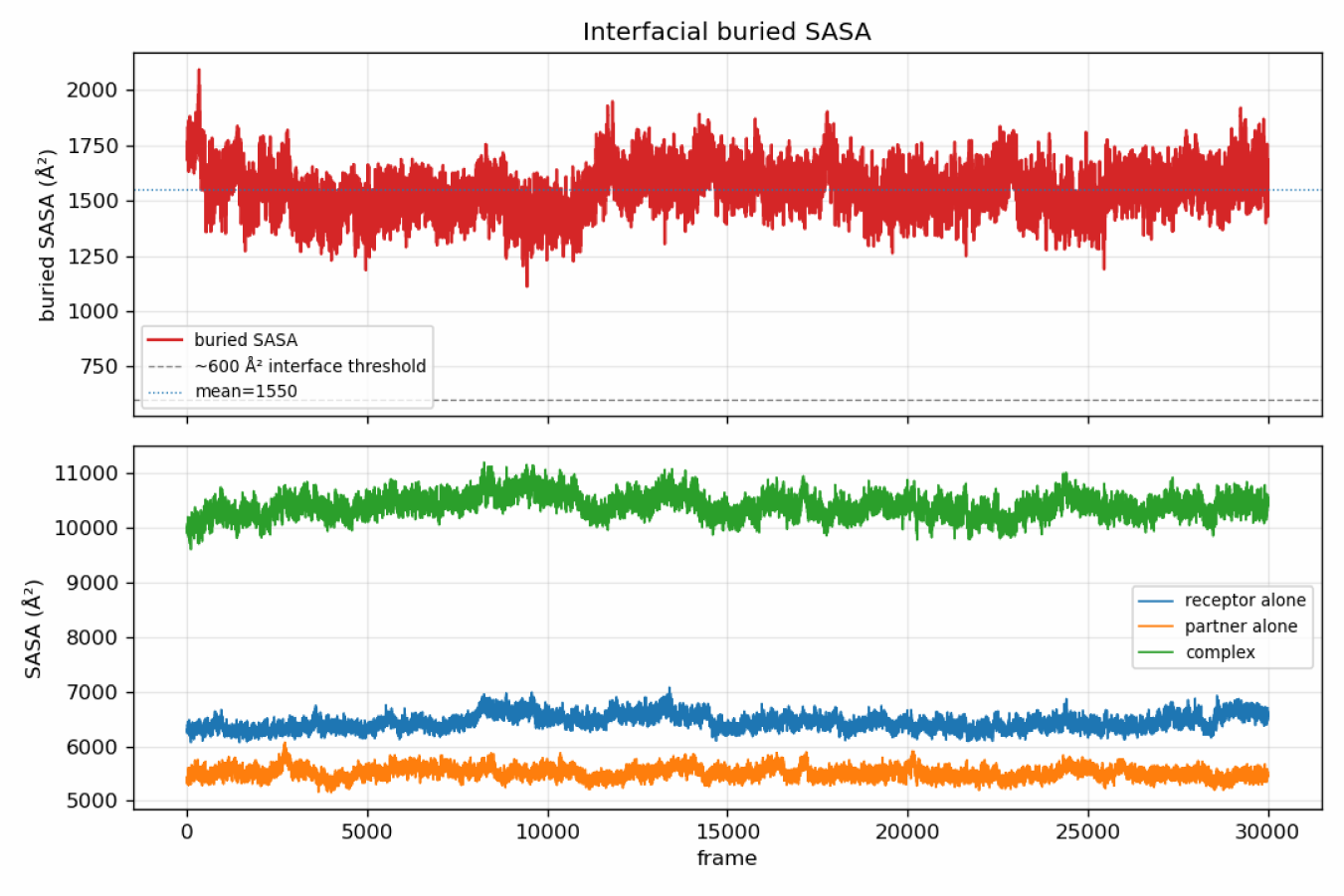}\caption{Buried interfacial SASA}\end{subfigure}\hfill
\begin{subfigure}[b]{0.45\linewidth}\includegraphics[width=\linewidth]{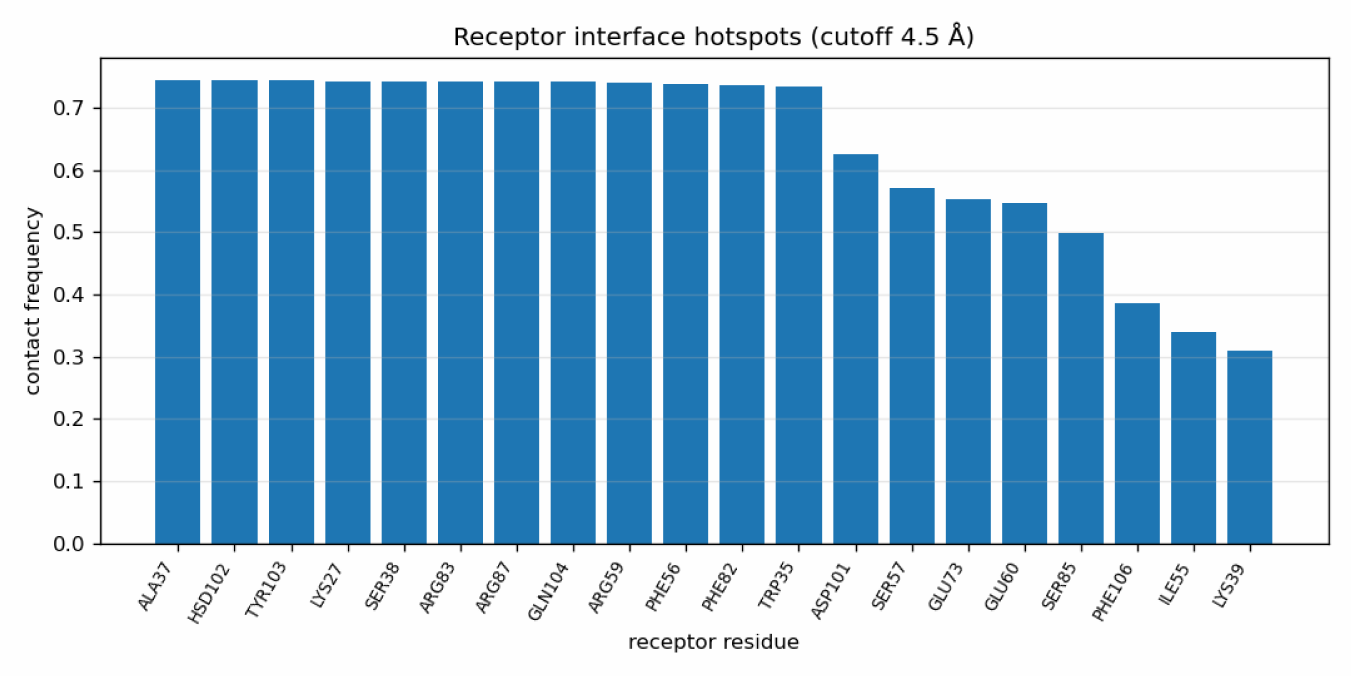}\caption{Interface contacts}\end{subfigure}
\caption{\footnotesize \textbf{Protein--protein (barnase--barstar, PDB 1BRS).} The partners stay folded and natively associated (per-chain backbone RMSD $\approx$1.3/0.85~\AA, interface RMSD $\approx$1.9~\AA), with a stable buried interface and retained hotspot contacts.}
\label{fig:si_bnbs}
\end{figure}

Fig.~S4a shows the per-chain backbone RMSD, low for both partners (1.27~\AA{} barnase, 0.85~\AA{} barstar), indicating that each protein remains individually folded throughout the run. Fig.~S4b gives the CAPRI-style interface RMSD (1.93~\AA), which stays in the near-native range for the full trajectory, showing the complex neither dissociates nor substantially rearranges. Fig.~S4c plots the buried interfacial SASA, approximately constant, showing the interface area is neither growing nor shrinking over time. Fig.~S4d shows the per-residue interface-contact frequency, with the native hotspot pairs (barnase Arg59/Arg83/Arg87/His102/Tyr103 with barstar Asp35/Asp39/Trp44) retained in the large majority of frames.

\FloatBarrier

\section{Agent prompts}
\label{si:prompts}

The framework is driven entirely by natural language: each case below was
started from a single high-level prompt, which the agent translated into the
ordered sequence of \texttt{htmd-mcp} tool calls (build, parametrize, minimize,
equilibrate, produce, analyze, compare). The prompts issued for the five case
studies are reproduced verbatim below.

\vspace{4pt}
\noindent\textbf{Case 1 --- Protein--DNA (p53 R248 hotspot mutants).}
\begin{lstlisting}[style=prompt,backgroundcolor=\color{p53bg},rulecolor=\color{p53frame}]
Using the htmd-mcp tools, build, simulate, and analyze the p53 DNA-binding
domain bound to its DNA response element from PDB 2AC0. Fetch and inspect the
structure; keep the two protein chains, the 12-bp DNA duplex, and the
structural Zn2+, and remove crystallographic waters. Prepare the protein at
pH 7.0 with PROPKA, optimize the H-bond network, rebuild missing atoms, and
auto-segment for psfgen. Build the solvated protein-nucleic system
(build_protein_nucleic) with CHARMM36 and TIP3P water, neutralized with
0.15 M KCl. Generate three variants under an identical protocol: wild type,
R248W, and R248G. For each variant, set up a restrained minimization
(10000 steps), an equilibration (NVT 0.5 ns then NPT 1 ns with a restraint
ramp on protein, nucleic, and Zn), and a 100 ns NPT production at a 2 fs
timestep (no HMR); then run NAMD3. After production, wrap_and_align each
trajectory and compute: backbone RMSD fit to the crystal structure with the
three terminal residues at each end excluded, per-residue RMSF, radius of
gyration, and the intermolecular protein-DNA hydrogen-bond count using a
GROMACS-style criterion (donor-acceptor <= 3.5 A, angle <= 30 deg). Use
compare_protein_nucleic to overlay WT/R248W/R248G and benchmark the trends
against Balasundaram and Priya Doss (2022).
\end{lstlisting}

\noindent\textbf{Case 2 --- Protein--membrane (M2 muscarinic receptor + iperoxo).}
\begin{lstlisting}[style=prompt,backgroundcolor=\color{membg},rulecolor=\color{memframe}]
Set up a membrane MD simulation of the active-state human M2 muscarinic
acetylcholine receptor bound to the agonist iperoxo, following the AMBER
Membrane-protein tutorial but using the available NAMD3/CHARMM htmd-mcp tools.
Fetch 4MQS OPM-aligned; keep chain A (the receptor) and discard the
T4-lysozyme fusion. Prepare the protein with PROPKA at pH 7.4, membrane-aware
(pass the OPM hydrophobic thickness). Extract iperoxo (ligand IXO) at its bound
pose, perceive bond orders and hydrogens, and parametrize it with SwissParam
(CGenFF). Embed the receptor in a POPC:cholesterol 9:1 bilayer (list_lipids,
then build_protein_membrane with an 80x80 A patch), solvate with TIP3P, and add
0.15 M NaCl. Run a minimization, a 2 ns restrained membrane-NPT equilibration
with a restraint ramp, and a 100 ns membrane-NPT production using
hydrogen-mass repartitioning at a 4 fs timestep; run NAMD3 in GPU-resident
mode. Analyze protein Ca RMSD, ligand heavy-atom RMSD (ligand_rmsd), RMSF,
radius of gyration, secondary structure, ligand-protein contacts, and hydrogen
bonds, and produce a QA stability dashboard. Confirm that iperoxo stays bound
through the canonical M2 orthosteric contacts (Asp103/D3.32, the tyrosine lid
Y104/Y403/Y426/Y430, Trp400/W6.48, and the Asn404/N6.52 hydrogen bond).
\end{lstlisting}
\newpage
\noindent\textbf{Case 3 --- Protein--ligand (congeneric TYK2 inhibitor series).}
\begin{lstlisting}[style=prompt,backgroundcolor=\color{ligbg},rulecolor=\color{ligframe}]
Run an end-to-end binding campaign for the congeneric TYK2 inhibitor series
using htmd-mcp. Starting from the TYK2 receptor (tyk2_protein.pdb) and the set
of 16 ligands (tyk2_ligands.sdf), for each ligand: standardize the ligand to a
parameter-ready mol2 (explicit H, 3D coordinates, charges) with
prepare_ligand_mol2 and parametrize it with SwissParam (CGenFF); seat it in the
orthosteric pose; and build the solvated protein-ligand complex
(build_protein_ligand) with CHARMM36/CGenFF, TIP3P water, and 0.15 M NaCl. Run
minimization, equilibration, and 10 ns of NPT production per complex, then run
NAMD3. For every complex compute ligand heavy-atom RMSD, protein RMSD, radius
of gyration, and the ligand-protein contact fingerprint, and estimate the
binding free energy with single-trajectory MM-GBSA (mm_pbsa). Collect all 16
Delta-G estimates, correlate them against the experimental affinities and the
Schrodinger FEP+ reference, and report Pearson and Spearman rank correlations
over the full 16-compound set (not a subset).
\end{lstlisting}

\noindent\textbf{Case 4 --- Protein--water (ubiquitin reference).}
\begin{lstlisting}[style=prompt,backgroundcolor=\color{watbg},rulecolor=\color{watframe}]
Build and simulate ubiquitin (PDB 1UBQ) as a solvated single-protein reference
using htmd-mcp. Fetch and inspect 1UBQ; prepare the protein with PROPKA at
pH 7.0; auto-segment for psfgen; and build a TIP3P water box
(build_protein_waterbox) with CHARMM36m, neutralized with 0.15 M KCl. Run a
restrained minimization, an NVT then NPT equilibration, and a multi-nanosecond
NPT production; then run NAMD3. After wrap_and_align, report Ca RMSD, radius of
gyration, secondary structure, SASA, and the hydrogen-bond count. Verify that
the beta-grasp fold is retained and that the radius of gyration matches the
crystallographic value (about 11.9-12.0 A). Report only rotationally-aligned
observables; if the trajectory was not rotationally aligned, omit the RMSF
rather than reporting whole-molecule tumbling.
\end{lstlisting}

\noindent\textbf{Case 5 --- Protein--protein (barnase--barstar complex).}
\begin{lstlisting}[style=prompt,backgroundcolor=\color{ppbg},rulecolor=\color{ppframe}]
Simulate the barnase-barstar complex (PDB 1BRS) with htmd-mcp and characterize
its interface. Fetch and inspect 1BRS; keep one barnase chain and one barstar
chain, and discard the extra crystallographic copies and waters. Prepare the
protein at pH 7.0 (PROPKA), auto-segment, and build the solvated
protein-protein complex (build_protein_protein_complex) with CHARMM36, TIP3P
water, and 0.15 M NaCl. Run minimization, equilibration, and about 300 ns of
NPT production, then run NAMD3. Analyze per-chain backbone RMSD (paper-matched),
CAPRI-style interface RMSD (interface_rmsd), per-residue interface contacts
(interface_contacts), and buried interfacial SASA (buried_sasa). Confirm the
partners stay natively associated and that the hotspot contacts (barnase
Arg59/Arg83/Arg87/His102/Tyr103 with barstar Asp35/Asp39/Trp44) are retained;
compare to Martin and Frezza (2022).
\end{lstlisting}

% =====================================================================
%  Tables S3 and S4 -- drop into the Supplementary Information,
%  after Table~\ref{tab:analysis} (the analysis-tool catalog).
%  They will auto-number as S3 and S4 under the existing
%  \renewcommand{\thetable}{S\arabic{table}} in the SI preamble.
%
%  Requires (all already in the manuscript preamble):
%    booktabs, threeparttable, siunitx-free -- no new packages needed.
% =====================================================================
\newpage

\section{Measured setup cost and error ledger}
\label{si:setupcost}
{Tables~\ref{tab:timing} and~\ref{tab:errors} provide the quantitative support
for the efficiency and error-reduction claims made in the Abstract and
Conclusions. Table~\ref{tab:timing} reports the instrumented wall-clock cost of
each pipeline stage; Table~\ref{tab:errors} is a complete ledger of every
failure and artifact logged across the five campaigns, together with how each
was detected and resolved.}

% ---------------------------------------------------------------------
%  TABLE S3 -- setup timing
% ---------------------------------------------------------------------
\begin{table}[htbp]
\centering
\begin{threeparttable}
\caption{{Measured wall-clock cost per pipeline stage for the congeneric TYK2
protein--ligand series. The agent-side setup path (ligand extraction through
solvated-system construction) completes in a median of 30~s per complex,
against $\sim$3~h of GPU production per complex. All stages were executed
sequentially on a single NVIDIA RTX~2080~Ti.}}
\label{tab:timing}
\small
\begin{tabular}{@{}p{3.3cm} p{4.9cm} r r r@{}}
\toprule
Stage & \texttt{htmd-mcp} tool(s) invoked & Median & Range & Total\tnote{a} \\
\midrule
Ligand extraction      & \texttt{fetch\_ligand\_from\_rcsb} / pose extraction & $<1$~s & $<1$~s & $<1$~s \\
Ligand standardization & \texttt{prepare\_ligand\_mol2}                       & $<1$~s & $<1$~s & $<1$~s \\
Parametrization        & \texttt{parametrize\_ligand} (SwissParam/CGenFF)     & 7~s    & 0--8~s\tnote{b} & 70~s \\
System construction    & \texttt{build\_protein\_ligand}                      & 23~s   & 23--25~s & 257~s \\
\addlinespace
\multicolumn{2}{@{}l}{\textbf{Agent-side setup subtotal}} & \textbf{30~s} & \textbf{24--32~s} & \textbf{327~s} \\
\midrule
Production MD (10~ns)  & \texttt{setup\_*}, \texttt{run\_namd3}               & 10\,733~s & 10\,700--10\,904~s & 118\,439~s \\
                       &                                                     & (2.98~h)  & & (32.9~h) \\
Trajectory analysis    & \texttt{wrap\_and\_align}, \texttt{ligand\_rmsd},    & 125~s  & 118--130~s & 1\,372~s \\
                       & \texttt{ligand\_protein\_contacts}, \texttt{compute\_rmsd} & & & \\
Binding energetics     & \texttt{mm\_pbsa}                                    & 108~s  & 102--190~s & 1\,385~s \\
\midrule
\multicolumn{2}{@{}l}{\textbf{Full pipeline, per complex}} & \textbf{3.06~h} & & \textbf{33.8~h} \\
\bottomrule
\end{tabular}
\begin{tablenotes}[flushleft]\footnotesize
\item[a] Totals are over the $n=11$ ligands processed by the instrumented
campaign runner. The remaining five compounds of the 16-ligand set
(\texttt{ejm\_43}, \texttt{ejm\_44}, \texttt{ejm\_47}, \texttt{jmc\_23},
\texttt{jmc\_28}) were run before instrumentation was added and are excluded
from the timing statistics only; they are included in all reported scientific
results.
\item[b] A 0~s parametrization is a hit on the local SwissParam result cache
for a previously parametrized ligand. One further ligand (\texttt{ejm\_49}) had
been pre-built by an earlier \texttt{stop-after=build} invocation, so its
recorded setup times are cache hits; its true values (7~s and 23~s) were
recovered from the campaign log and used here.
\item[] Source data: \texttt{p\_l/tyk2/summary/campaign\_status.json} and
\texttt{campaign.log}.
\end{tablenotes}
\end{threeparttable}
\end{table}

% ---------------------------------------------------------------------
%  TABLE S4 -- error ledger
% ---------------------------------------------------------------------
\begin{table}[htbp]
\centering
\begin{threeparttable}
\caption{{Complete ledger of failures and artifacts logged across the five
campaigns, with detection mechanism and resolution. Infrastructure failures
were recovered without loss of any scientific result; the parametrization
artifacts were corrected at source in the \texttt{htmd-mcp} server; the
analysis artifacts were detected by the agent and either corrected or withheld
rather than reported.}}
\label{tab:errors}
\small
\begin{tabular}{@{}p{2.7cm} p{4.3cm} p{3.4cm} p{3.9cm} p{2.5cm}@{}}
\toprule
System / stage & Failure mode & Detection & Resolution & Outcome \\
\midrule
\multicolumn{5}{@{}l}{\textit{Infrastructure and build-time failures} (9 logged events, 4 root causes)} \\
\addlinespace
Protein--ligand, \texttt{mol2} & \texttt{ModuleNotFoundError: htmd\_mcp} --- package not on the subprocess \texttt{PYTHONPATH} & Runner exception trace & Environment corrected, batch restarted & Recovered in 43~s \\
\addlinespace
Protein--ligand, \texttt{parametrize} ($\times 3$) & \texttt{URLError}: name resolution failure reaching the SwissParam service & \texttt{urllib} exception in runner log & Re-driven once network access was restored & All 3 completed \\
\addlinespace
Protein--ligand, \texttt{parametrize} ($\times 3$) & Truncated cache entry from the interrupted download (\texttt{MATCH/} folder missing in cached result) & Cache-integrity check in \texttt{parametrize\_ligand} & Poisoned cache entries evicted and re-fetched & All 3 completed \\
\addlinespace
Protein--ligand, \texttt{parametrize} ($\times 2$) & Repeat name-resolution failure on the same service & As above & Retried & Both completed \\
\midrule
\multicolumn{5}{@{}l}{\textit{Force-field artifacts, corrected at source}} \\
\addlinespace
Protein--membrane, ligand parametrization & MATCH copied triple-bond parameters onto two C(sp)--C(sp$^3$) single bonds flanking the iperoxo alkyne ($r_0 = 1.22$~\AA, catastrophically stiff $K_b$) & Automated artifact detector in \texttt{parametrize\_ligand} (\texttt{swissparam.py}) & $r_0$ restored to $\approx 1.50$~\AA; general fix keyed on bonds flanking a triple bond & Corrected before production \\
\addlinespace
Protein--membrane, ligand parametrization & Seven dihedral terms routed through the linear sp alkyne centres retained non-zero force constants, where CGenFF assigns zero & Near-linear angle detection ($\theta_0 \geq 170^{\circ}$) in the same detector & Force constants zeroed to match CGenFF & Two equilibration crashes eliminated \\
\midrule
\multicolumn{5}{@{}l}{\textit{Analysis artifacts detected by the agent}} \\
\addlinespace
Protein--water, analysis & RMSF computed on a trajectory that had not been rotationally aligned, reporting whole-molecule tumbling rather than internal flexibility & Agent inspection of the analysis provenance & Observable withheld rather than reported & RMSF omitted (Fig.~\ref{fig:si_ubq}) \\
\addlinespace
Protein--DNA, analysis & Default RMSD fit (production frame~0, mobile termini retained) inverted the WT\,/\,mutant ordering after $\sim$80~ns & Inconsistency with the H-bond and RMSF orderings & Refit to the crystal structure with three terminal residues excluded at each end & Expected ordering restored \\
\addlinespace
Protein--ligand, analysis & Apparent five-ligand MM-GBSA correlation ($r = 0.81$) not reproduced on the full congeneric set & Extension of the ranking to all 16 compounds & Small-sample artifact identified and reported as such & Negative result reported ($r = 0.31$) \\
\midrule
\multicolumn{5}{@{}l}{\textit{Unrecovered}} \\
\addlinespace
Protein--DNA, R248Q production & External termination (\texttt{SIGTERM}, \texttt{rc}$=-15$) at 44.6 of 100~ns; energies stable to the point of termination & Orchestrator return code & Not resumed & Variant excluded from the reported set\tnote{a} \\
\bottomrule
\end{tabular}
\begin{tablenotes}[flushleft]\footnotesize
\item[a] Six p53 variants were built (WT, R248W, R248G, R248Q, R248L, R248P),
four were carried to production, and the three completing the full 100~ns
protocol (WT, R248W, R248G) are those reported in the main text.
\item[] Source data: \texttt{p\_l/tyk2/summary/campaign.log},
\texttt{p53\_R248\_md/orchestrator.log}, and the M2 reproducibility report.
\end{tablenotes}
\end{threeparttable}
\end{table}

\clearpage
\newpage
\section{Cross-model orchestration comparison (protein--DNA)}
\label{si:llm}

{The three LLM orchestrators compared in the main text each built the mutants,
solvated, minimised, equilibrated, ran 100~ns of production, and analysed the
resulting trajectories. Nothing was shared between the three runs except the
starting PDB entry and the htmd-mcp server tool surface. Table~S6
reports the framework-level comparison behind that result.
As the table shows, Nemotron 3 Ultra consumed roughly $90\times$ the tokens
GPT-5.6 Sol did for the identical three simulations. This is not because it
reasons more: its output is only $11\times$ GPT's. It is because it issued
$24\times$ as many model calls, each carrying $3.8\times$ more context. Claude
Opus 4.8 sits between the two and is closer to Nemotron in shape --- many turns,
each moderately sized.
The three headline cost figures are not on a single scale. GPT's \$6.10 is an
API-equivalent estimate for a subscription session; Nemotron's \$0.00 is a real
recorded charge on a free endpoint; Claude's range is a projection at list
rates, because the transcripts of its campaign were deleted before the token
counters could be read. Normalised to a common Opus price, the ordering reverses
completely: GPT $\approx$\$6, Claude $\approx$\$42--66, and Nemotron \$637. The
free endpoint is by far the most expensive workload; it simply is not billed.
Per simulated system that is \$2, \$14--22, and \$212 respectively.
What drives the bill is context, not reasoning, in all three frameworks:
93--98\% of every model's input consisted of cache reads, and output tokens
account for under 0.5\% of total traffic everywhere. Nemotron's \$637 breaks
down as \$361 of cache reads plus \$268 of uncached input --- its 6.9\% uncached
fraction is twice GPT's 2.5\%, and that is where the real cost is incurred.}

{\footnotesize
\renewcommand{\arraystretch}{1.2}
\setlength{\LTcapwidth}{\textwidth}
\begin{longtable}{@{}p{3.4cm} p{3.5cm} p{3.5cm} p{4.1cm}@{}}
\caption{{Framework-level comparison of the three LLM orchestrators driving the
identical p53 DBD--DNA protocol. Values given as ranges for Claude Opus 4.8 are
estimates, not measurements: the three Claude Code session transcripts were
deleted, and token usage is recorded only there. They are scaled from two intact
htmd-mcp Claude Code sessions on the same machine (2.65 and 4.69~M tokens per
top-level prompt) applied to the campaign's 17 recovered task prompts. Entries
marked ``partial'' were captured before the run ended: the GPT-5.6 snapshot was
taken 31~minutes into a campaign that continued for a further six days of
simulation monitoring, and one Nemotron assistant record lacks a completion
timestamp.}}
\label{tab:llm_framework} \\

\toprule
\textbf{Category} & \textbf{Claude Opus 4.8} & \textbf{GPT-5.6 Sol}
& \textbf{Nemotron 3 Ultra} \\
\midrule
\endfirsthead

\multicolumn{4}{@{}l}{\textit{Table S6, continued}} \\
\toprule
\textbf{Category} & \textbf{Claude Opus 4.8} & \textbf{GPT-5.6 Sol}
& \textbf{Nemotron 3 Ultra} \\
\midrule
\endhead

\midrule
\multicolumn{4}{r@{}}{\textit{continued on the next page}} \\
\endfoot

\bottomrule
\multicolumn{4}{@{}p{14.9cm}@{}}{\footnotesize
$^{\mathrm{a}}$~Estimated, not measured; see the caption.\newline
$^{\mathrm{b}}$~Recorded charge on the OpenCode free endpoint. Priced at the
Opus-tier list rates in this table, the same traffic would come to \$637:
\$268 of uncached input, \$361 of cache reads and \$8 of output. The zero should
not be read as a paid NVIDIA NIM or enterprise price.} \\
\endlastfoot

\multicolumn{4}{@{}l}{\textit{Identity}} \\
\addlinespace
Provider & Anthropic & OpenAI & OpenCode routing, NVIDIA upstream \\
Model ID & \texttt{claude-opus-4-8} & \texttt{gpt-5.6-sol} & \texttt{nemotron-3-ultra-free} \\
Harness & Claude Code CLI + htmd-mcp & Codex CLI 0.148.0 + htmd-mcp & OpenCode CLI 1.18.22 + htmd-md MCP \\
Session record & 3 sessions: 19a82158, b723b27f, c8bd3391 --- transcripts deleted & Codex rollout, snapshot only & \texttt{ses\_fca2abeb6ffeuIPvhqrobppTb6} \\
Run window & 2026-05-19 $\rightarrow$ 2026-05-29 & 2026-08-20 $\rightarrow$ 2026-08-26 & 2026-08-24 $\rightarrow$ 2026-08-27 \\
\midrule

\multicolumn{4}{@{}l}{\textit{Model parameters}} \\
\addlinespace
Platform system prompt & Not exportable (harness) & Not exportable (OpenAI platform) & Not stored or exportable \\
Disclosed task-level system prompt & None --- free-form prompts; all 17 recovered & Yes; reproduced in Section~\ref{si:prompts} & None --- user prompt only; reproduced in Section~\ref{si:prompts} \\
Sampling temperature & Rejected by the API on this model: \texttt{temperature}, \texttt{top\_p} and \texttt{top\_k} were removed on Opus 4.8, and sending one returns HTTP 400 & Not configured or exposed by the Codex rollout & Supported, but no value configured or recorded \\
Reasoning / thinking & Adaptive thinking; \texttt{budget\_tokens} removed on this model & \texttt{reasoning\_effort=high}, standard mode & Enabled; no effort level or token budget recorded \\
Effort level & Claude Code default \texttt{xhigh} & High & Not recorded \\
Context window & 1~M & Short-context tier; no request exceeded the 272~K threshold & Not recorded \\
Service tier & Not recorded & Default & Not recorded \\
List price, input / output per M tokens & \$5.00 / \$25.00 & \$5.00 / \$30.00 & \$0.00 / \$0.00, free endpoint \\
List price, cache read / write per M tokens & \$0.50 / \$6.25 & \$0.50 / n/a & \$0.00 / \$0.00 \\
\midrule

\multicolumn{4}{@{}l}{\textit{Prompts and invocations}} \\
\addlinespace
Top-level task prompts & 17 & 1 & 1 + 6 human follow-ups \\
Model invocations & $\sim$340--400$^{\mathrm{a}}$ & 73, partial & 1{,}734 successful of 2{,}052 stream attempts \\
Failed / retried streams & Not recoverable & Not recorded & 312; $287 \times$ NVIDIA 502 overload, $17 \times$ 404 \\
\midrule

\multicolumn{4}{@{}l}{\textit{Tokens}} \\
\addlinespace
Uncached input tokens & 1.6--2.9~M$^{\mathrm{a}}$ & 215{,}255 & 53{,}545{,}692 \\
Cached input tokens & 43--76~M$^{\mathrm{a}}$ & 8{,}415{,}488 & 722{,}705{,}760 \\
Cache-write tokens & 0.3--0.6~M$^{\mathrm{a}}$ & 0 & 0 \\
Cache fraction of input & 95.5--97.3\%$^{\mathrm{a}}$ & 97.5\% & 93.10\% \\
Output tokens, incl. reasoning & 0.2--0.4~M$^{\mathrm{a}}$ & 27{,}335; 11{,}937 reasoning & 309{,}122; 32{,}402 reasoning \\
Total tokens & 45--80~M$^{\mathrm{a}}$ & 8{,}658{,}078 & 776{,}560{,}574 \\
Mean tokens / invocation & $\sim$130{,}000--200{,}000$^{\mathrm{a}}$ & 118{,}604 & 447{,}843 \\
Mean tokens / top-level prompt & 2.65--4.69~M$^{\mathrm{a}}$ & 8{,}658{,}078 & 776{,}560{,}574 \\
\midrule

\multicolumn{4}{@{}l}{\textit{Cost}} \\
\addlinespace
API-equivalent cost & \$42--\$66$^{\mathrm{a}}$ & \$6.10 & \$0.00 recorded; free endpoint$^{\mathrm{b}}$ \\
Cost per top-level prompt & \$2.45--\$3.88$^{\mathrm{a}}$ & \$6.10 & \$0.00$^{\mathrm{b}}$ \\
\midrule

\multicolumn{4}{@{}l}{\textit{Time}} \\
\addlinespace
LLM wall time & Not recoverable & 1{,}879~s, partial & 152{,}406~s $=$ 42~h 20~min, partial \\
Session span & 9.7 days & $\sim$6 days & 68~h 34~min \\

\end{longtable}
}

\end{document}